\documentclass[runningheads]{llncs}
\usepackage{versions}\excludeversion{ignore}
\usepackage{amsmath}
\DeclareMathOperator*{\argmax}{argmax}
\usepackage{amssymb}
\usepackage{graphicx}
\usepackage{csquotes}\MakeOuterQuote{"}
\usepackage{todonotes,soul}
\usepackage{subcaption}
\usepackage[inline]{enumitem}
\usepackage{booktabs} 
\usepackage{xcolor}
\usepackage{multirow}
\usepackage[T1]{fontenc}
\usepackage{pgfplots} 
\usepackage{array}
\usepackage{xspace}

\newcommand{\layoutfont}{\fontfamily{Cabin-TLF}\selectfont}
\newcommand{\layoutglyphnospace}[1]{\mbox{{\layoutfont #1}}}
\newcommand{\layoutglyph}[1]{\layoutglyphnospace{#1}\xspace}

\newcommand{\CC}{\layoutglyph{C}}
\newcommand{\LL}{\layoutglyph{L}}
\newcommand{\UU}{\layoutglyph{U}}
\newcommand{\OO}{\layoutglyph{O}}
\newcommand{\YY}{\layoutglyph{Y}}
\newcommand{\CR}{\reflectbox{\layoutglyphnospace{C}}\xspace}
\newcommand{\LR}{\reflectbox{\layoutglyphnospace{L}}\xspace}
\newcommand{\UR}{\raisebox{7pt}{\scalebox{1}[-1]{\layoutglyphnospace{U}}}\xspace}

\newcommand{\rev}[1]{#1} %{\textcolor{blue}{#1}}
\usepackage[hidelinks]{hyperref}

\usepackage{orcidlink}
\renewcommand{\orcidID}[1]{\orcidlink{#1}}
\begin{document}
\title{Complex Layout Classification in the Wild:\\ A Low-Resource Approach with Layout-Preserving Augmentations
}

\author{Sharva Gogawale\orcidID{0009-0000-5230-5197} \and
Iddo Hakim%\orcidID{0000-0000-0000-0000} 
\and
Gal Grudka
%\orcidID{0000-0000-0000-0000} 
\and
Mohammad Suliman
\orcidID{0009-0007-5550-3448} 
\and
Omer Ventura%\orcidID{0000-0000-0000-0000} 
\and
Daria Vasyutinsky-Shapira\orcidID{0000-0002-4257-7882} \and
Berat Kurar-Barakat\orcidID{0000-0002-7240-7286} \and ~~~~~~~
Nachum Dershowitz
%\orcidID{0000-0000-0000-0000}
}

\titlerunning{Complex Layout Classification in the Wild}
\authorrunning{S. Gogawale et al.}

\institute{School of Computer Science and AI, Tel Aviv University, Ramat Aviv, Israel\\
\email{\{sharvag,iddoh,galgrudka,suliman,omerventura\}@mail.tau.ac.il}\\
\email{\{dariashap,berat,nachumd\}@tauex.tau.ac.il}}

\sloppy %%%%
\maketitle   
%\vspace*{-2em}
% typeset the header of the contribution
%
\begin{abstract}
%Document layout analysis is crucial for digitization of historical collections. But 
Many digitized corpora suffer from low resources because annotations may be scarce, page scans are noisy and of poor resolution, or layouts are structurally complex in ways that negatively affect the quality of automatic transcription. 
Developing robust classification models for low-resource languages is inhibited by the lack of large-scale annotated data and by the frequent semantic complexity of page layouts. 
To this end, we have curated a complex-layout dataset, 
%a compact corpus of 155 high-resolution page images drawn from the National Library of Israel’s digital collections of printed books, 
manually classified into eight distinct layout types based on their separator regions. 
To overcome data scarcity, we propose a novel training strategy in the form of a CNN-based classifier that employs strong, domain-aware augmentations to improve generalization. 
We utilize narrow anisotropic Gaussian masking to suppress incidental textual details while preserving essential separations, compelling the model to learn global geometric arrangements. 
Additionally, we implement reflection-induced label transformations to enrich the training distribution while maintaining label consistency across asymmetric categories.
\rev{The results demonstrate that layout-specific augmentations can substantially improve page-level layout classification under severe annotation scarcity.}
% The approach demonstrates that utilizing layout-specific augmentations enables robust learning even with minimal annotated samples. 
%We hope that this work will provide a foundational step toward improving downstream transcription systems by effectively routing pages to layout-aware pipelines. 
%The dataset and source code can be accessed at [Anonymized link].
\end{abstract}

\section{Introduction}

Document layout analysis is a fundamental task in document understanding, which serves as a critical prerequisite for downstream pipelines, such as optical character recognition (OCR),  automated transcription~\cite{dla},  entity extraction, and digital reconstruction of reading orders. 
Within layout analysis, layout classification refers to the task of assigning the layout of a page to a specific category. 
This is especially useful for historical documents, as it allows each page to be directed to the best processing pipeline. 
Then, later steps (e.g.\@ segmentation, OCR, and reading-order reconstruction) can be adapted to the specific layout. 

Modern documents such as digital PDFs, invoices, or scientific papers usually follow clean and well-organized grid structures, often called "Manhattan" layouts. In contrast, historical documents are much more irregular and complex, and often contain a high level of visual noise. 
This makes layout classification not merely a visual segmentation task but an important preliminary step in separating the logical structures from physical attributes. Correspondingly, layout analysis can be broadly categorized into visual-based physical layout analysis and semantics-based logical analysis~\cite{dla2}. Visual-based analysis relies on appearance and geometric cues to differentiate regions with different visual attributes, whereas logical analysis builds upon this segmentation by organizing regions according to the semantic role and relationships expressed in the text~\cite{dla3}.

\begin{figure}[t]
    \centering
\begin{subfigure}[b]{0.35\textwidth}
        \centering
        \includegraphics[width=\linewidth,trim=230 55 25 65,clip]{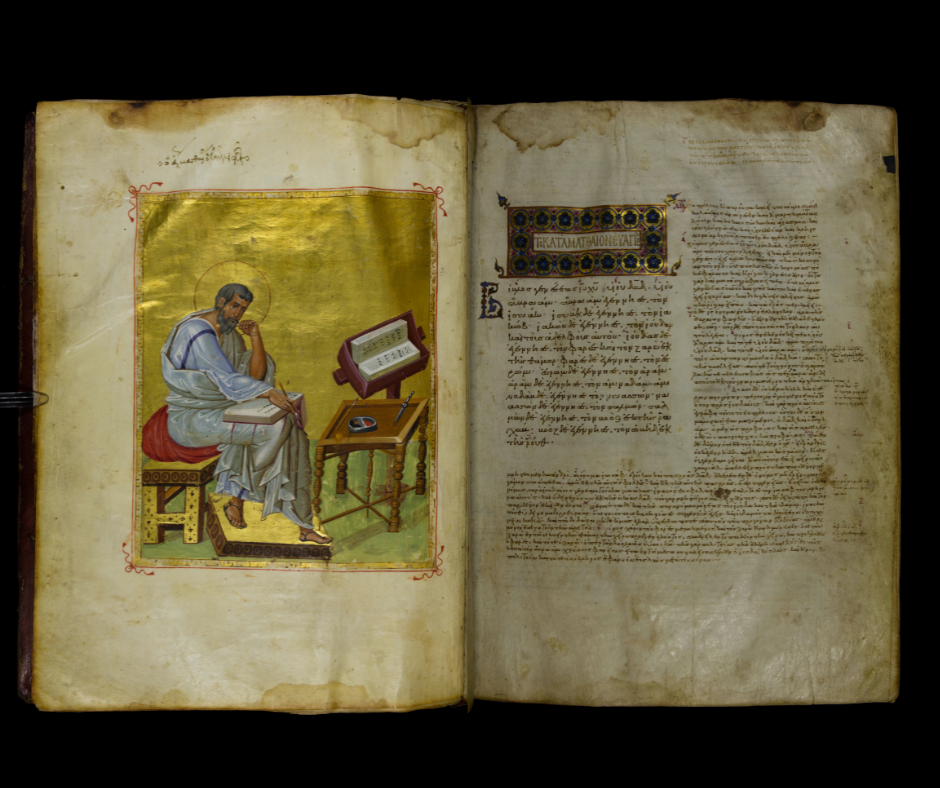}
    %    \caption{Greek Gospels with catena commentary, illuminated parchment (10th c.\@ Byzantium). National Library of Greece.}
    %    \label{fig:sub1}
    \end{subfigure}\qquad
    \begin{subfigure}[b]{0.35\textwidth}
        \centering
        \includegraphics[width=0.74\linewidth,trim=110 120 100 200,clip]{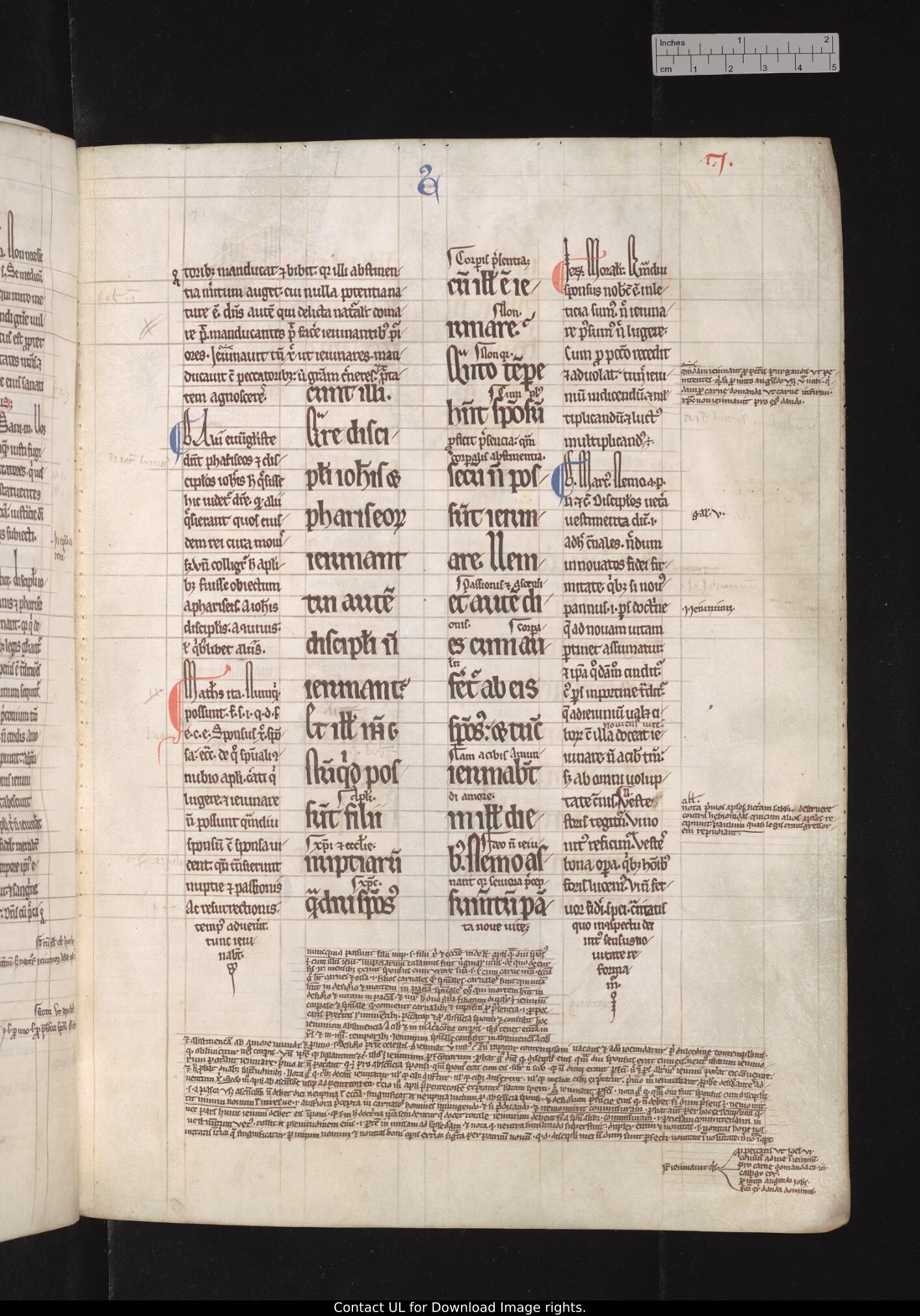}
 %       \caption{ Gospels of Mark, Luke, and John, with glosses (first half of the 13th century). Cambridge, Queens' College, MS 26, 4r.}
 %       \label{fig:sub2}
    \end{subfigure}
    \caption{Sample parchment pages with complicated layouts. Left:
    Greek Gospels with catena (10th c.\@ Byzantium), National Library of Greece. Right:
   Gospels in Latin  with glosses (England, 1300--1350), Cambridge, Queens' College, MS 26.}
   %, 4r.}
    \label{Fig1}
\end{figure}

Complex layouts (Fig.~\ref{Fig1}) are often found in medieval manuscripts before the era of print (until the mid-16th century), and are present in recently printed books, too. 
% Developing robust models for classification of such layouts will dramatically improve the quality of OCR for both manuscripts and printed books. 
\rev{Developing robust models for classifying such layouts can support layout-aware segmentation and OCR pipelines for both manuscripts and printed books.}
Because of the uniformity of printed scripts, it is easier to first solve this problem for printed books, and then proceed to manuscripts. 

\begin{figure}[t]
    \centering
        \begin{subfigure}{0.2\linewidth}
        \centering
        \includegraphics[ height=24 mm]{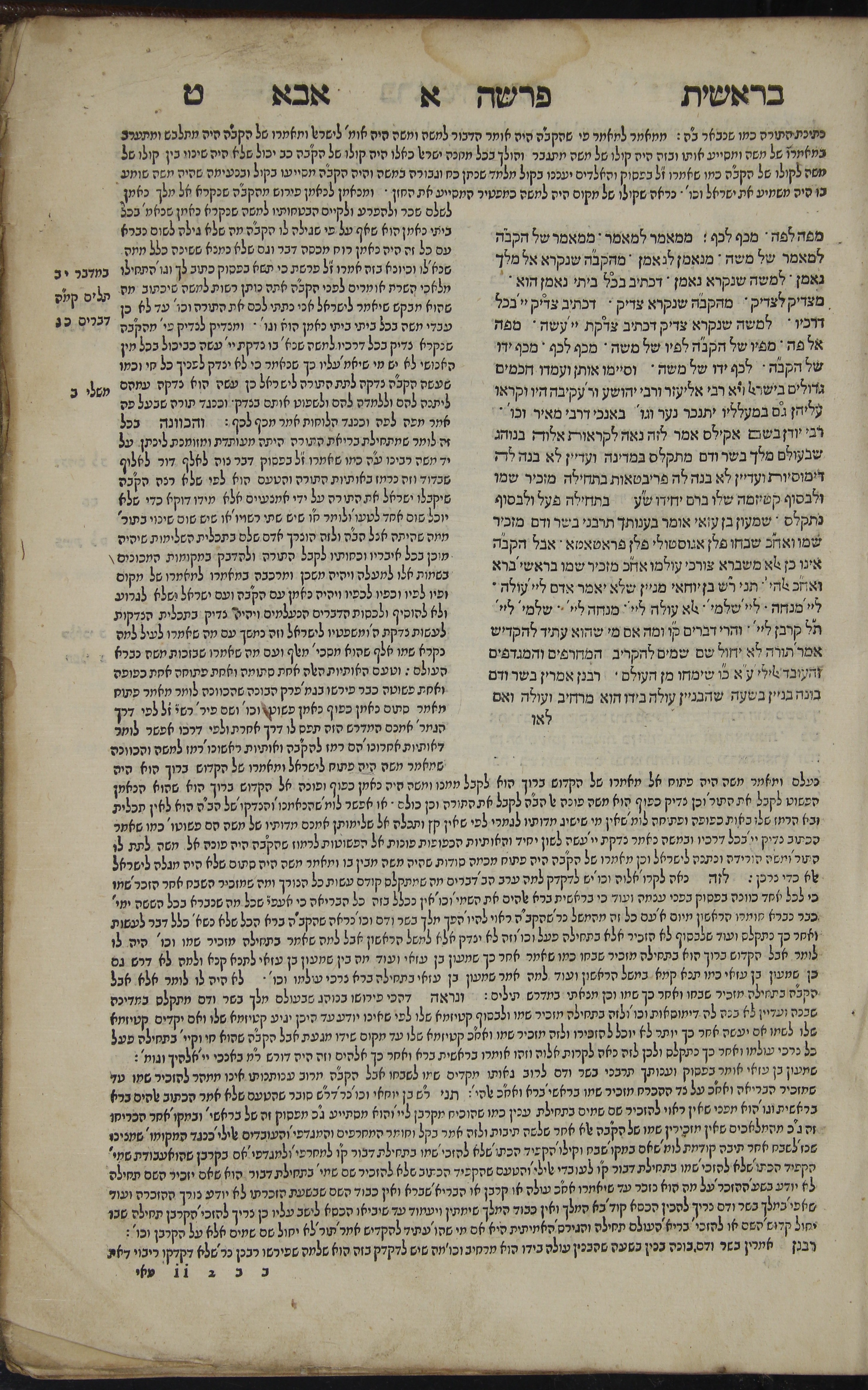} 
        \caption*{\CC}
        \label{fig:sub-c}
    \end{subfigure}
    \hfill
    \begin{subfigure}{0.2\linewidth}
        \centering
        \includegraphics[ height=24 mm]{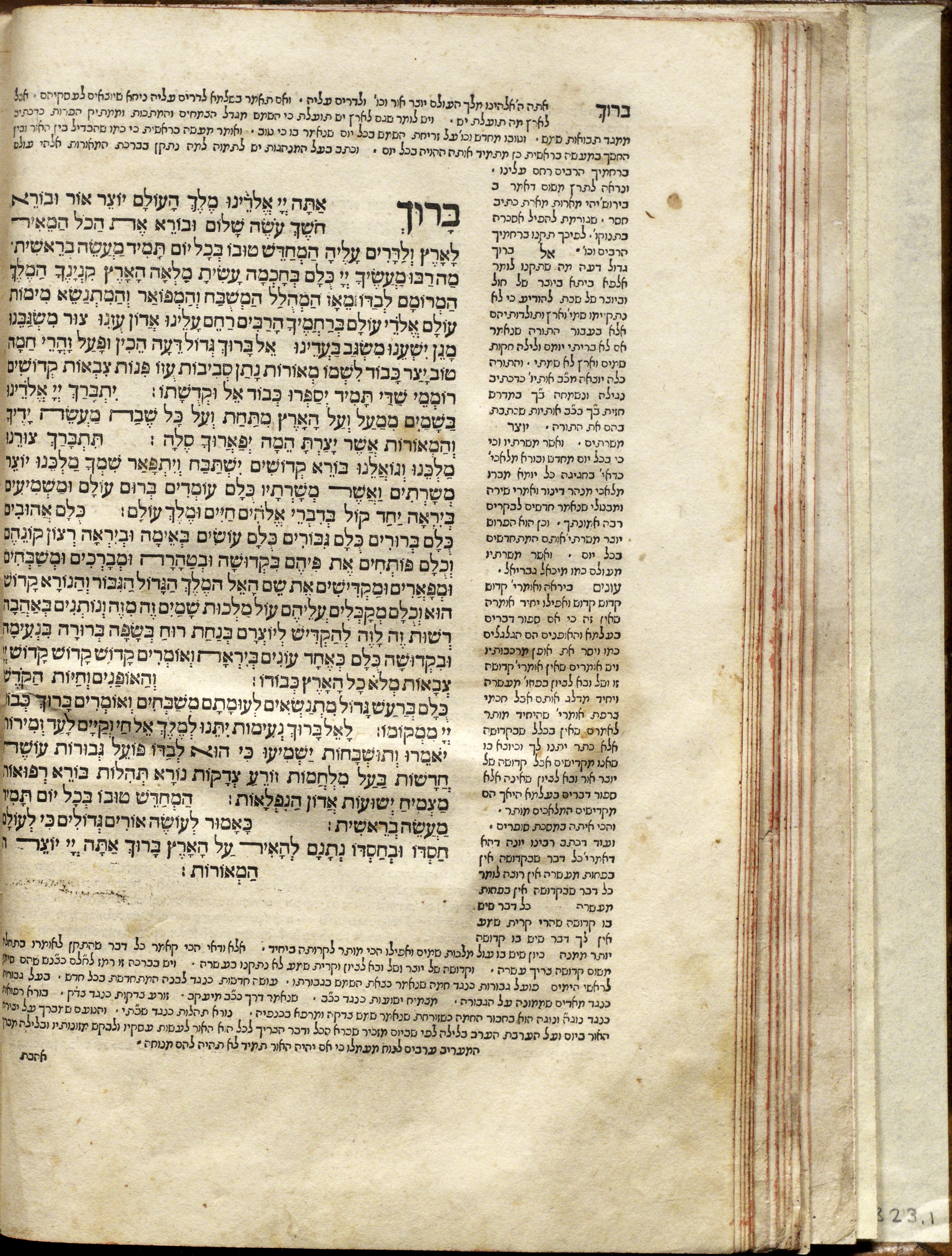} 
        \caption*{\CR}
        \label{fig:sub-c2}
    \end{subfigure}
    \hfill
    \begin{subfigure}{0.2\linewidth}
        \centering
        \includegraphics[ height=24 mm]{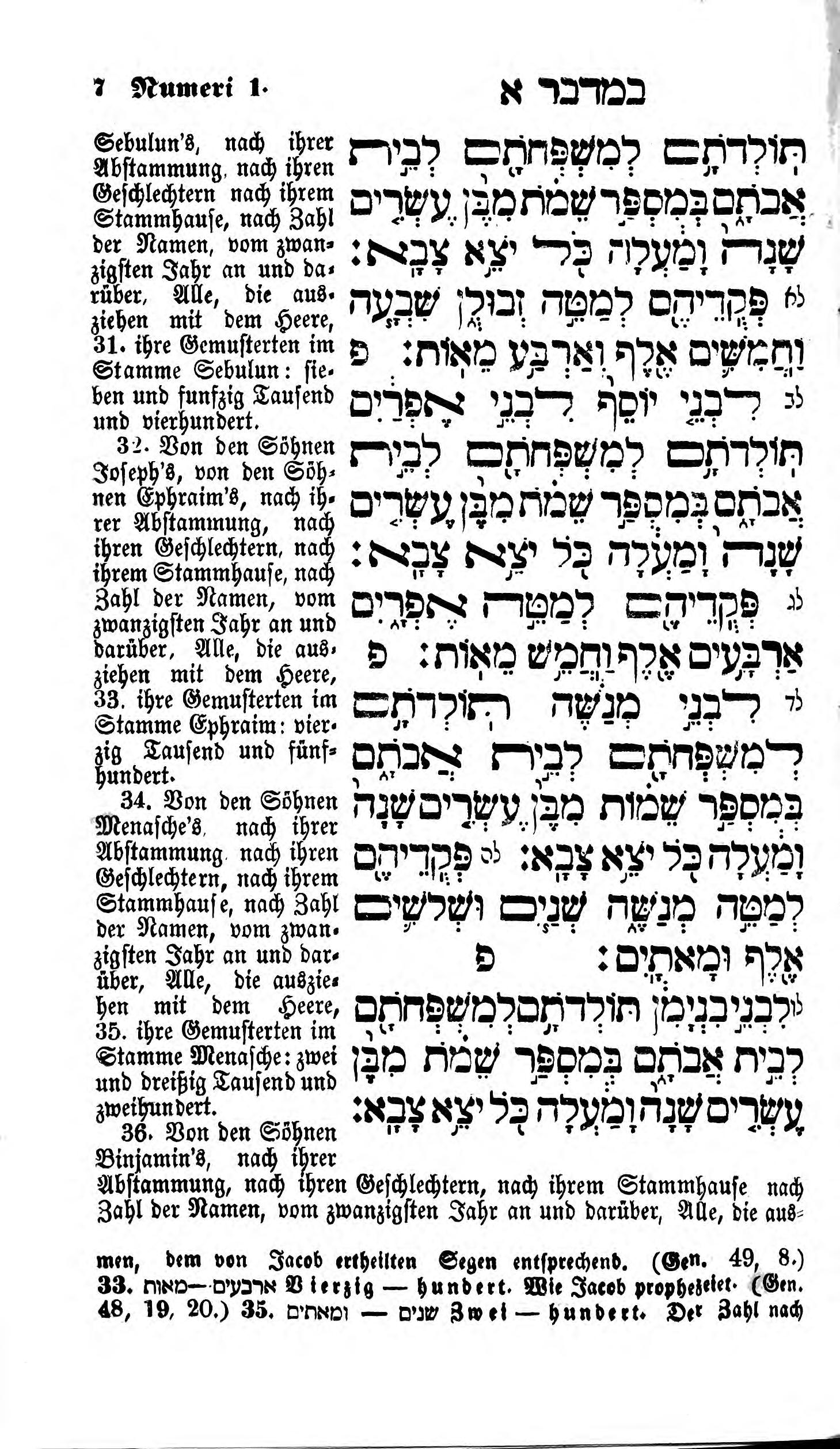} % 
        \caption*{\LL}
        \label{fig:sub-l}
    \end{subfigure}
    \hfill
    \begin{subfigure}{0.2\linewidth}
        \centering
        \includegraphics[ height=24 mm]{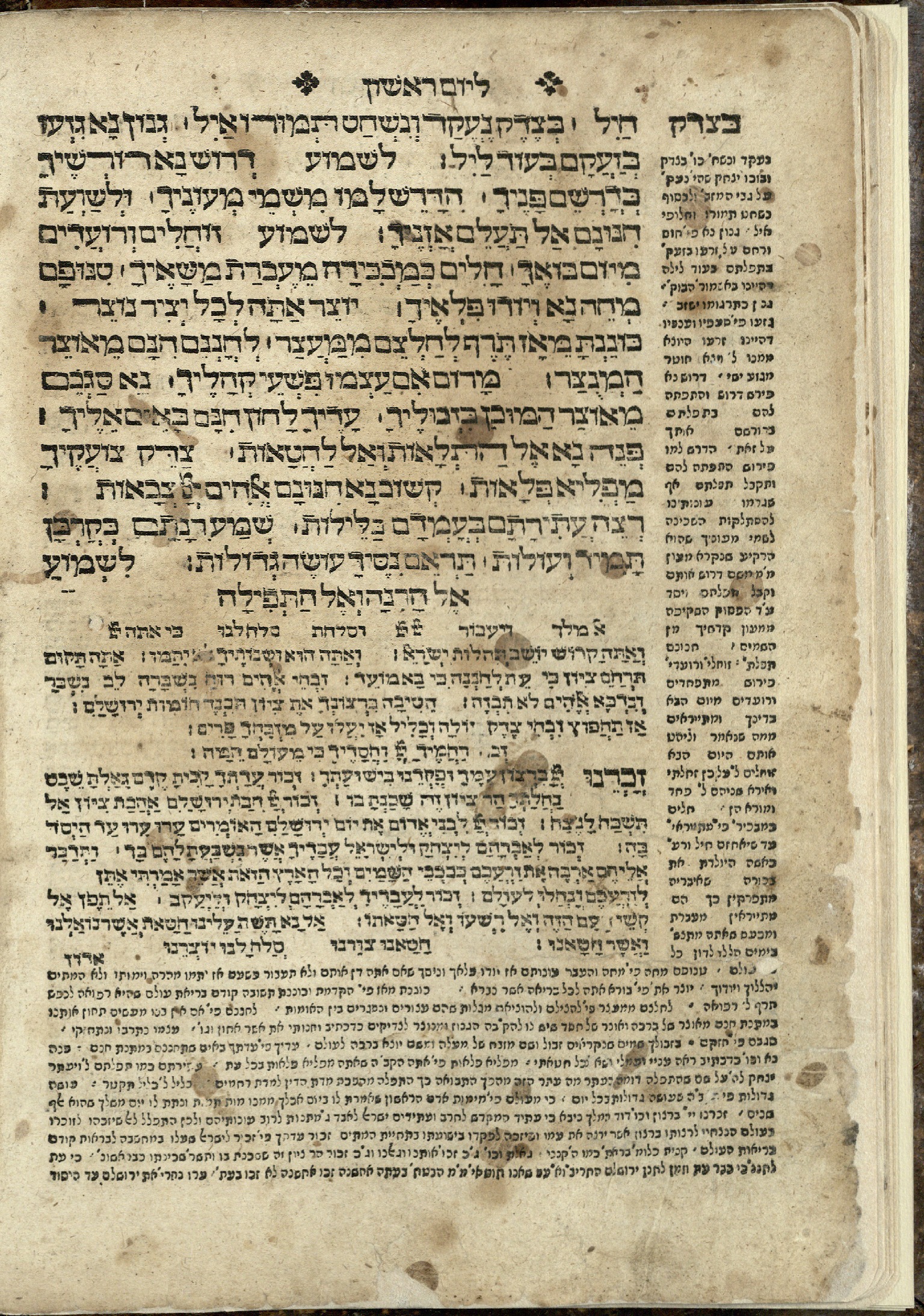} 
        \caption*{\LR}
        \label{fig:sub-l2}
    \end{subfigure}
    
    \vspace{0.2em} 

    \begin{subfigure}{0.2\linewidth}
        \centering
        \includegraphics[ height=24 mm]{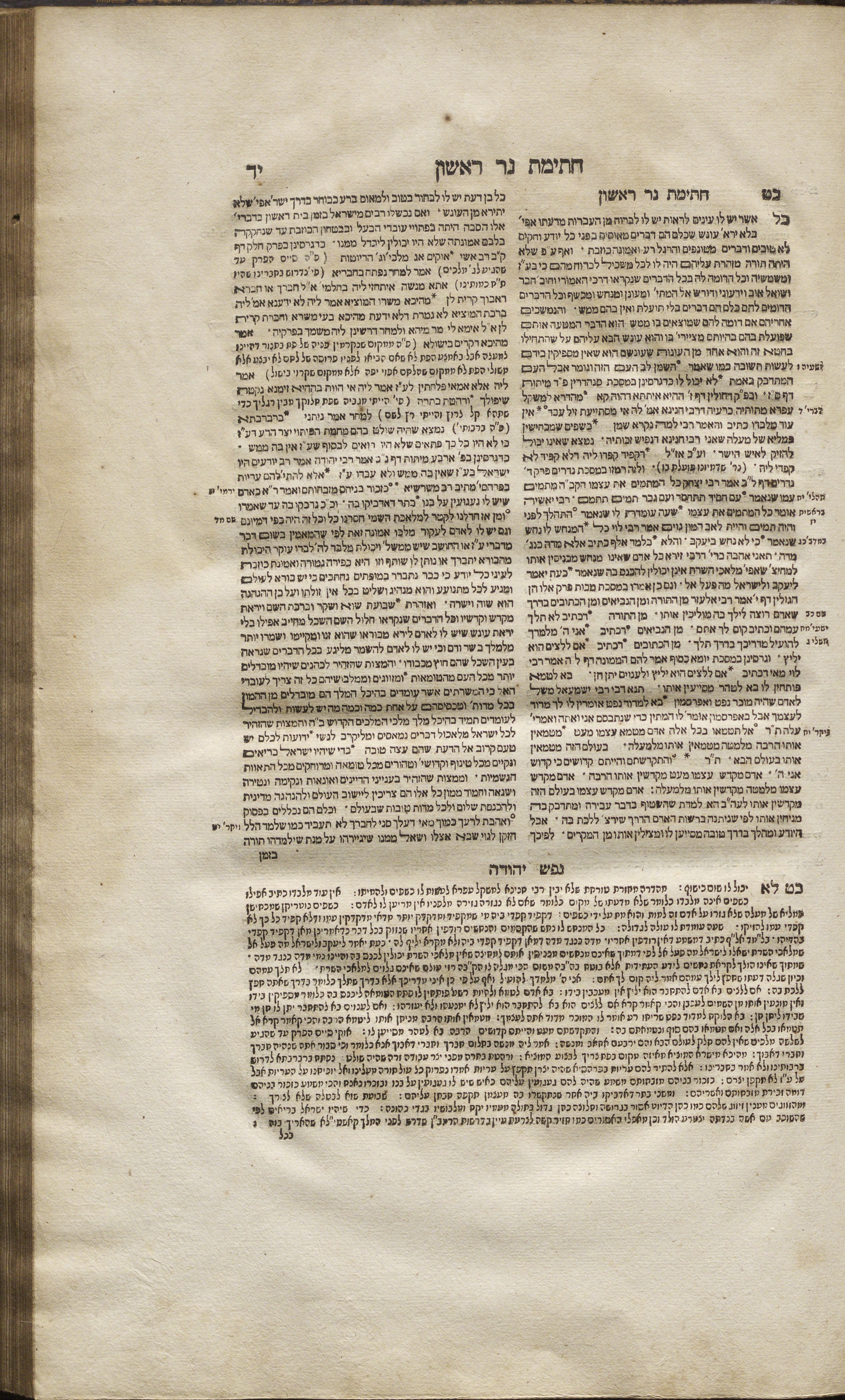} 
        \caption*{\UU}
        \label{fig:sub-u}
    \end{subfigure}
    \hfill
    \begin{subfigure}{0.2\linewidth}
        \centering
        \includegraphics[height=24 mm]{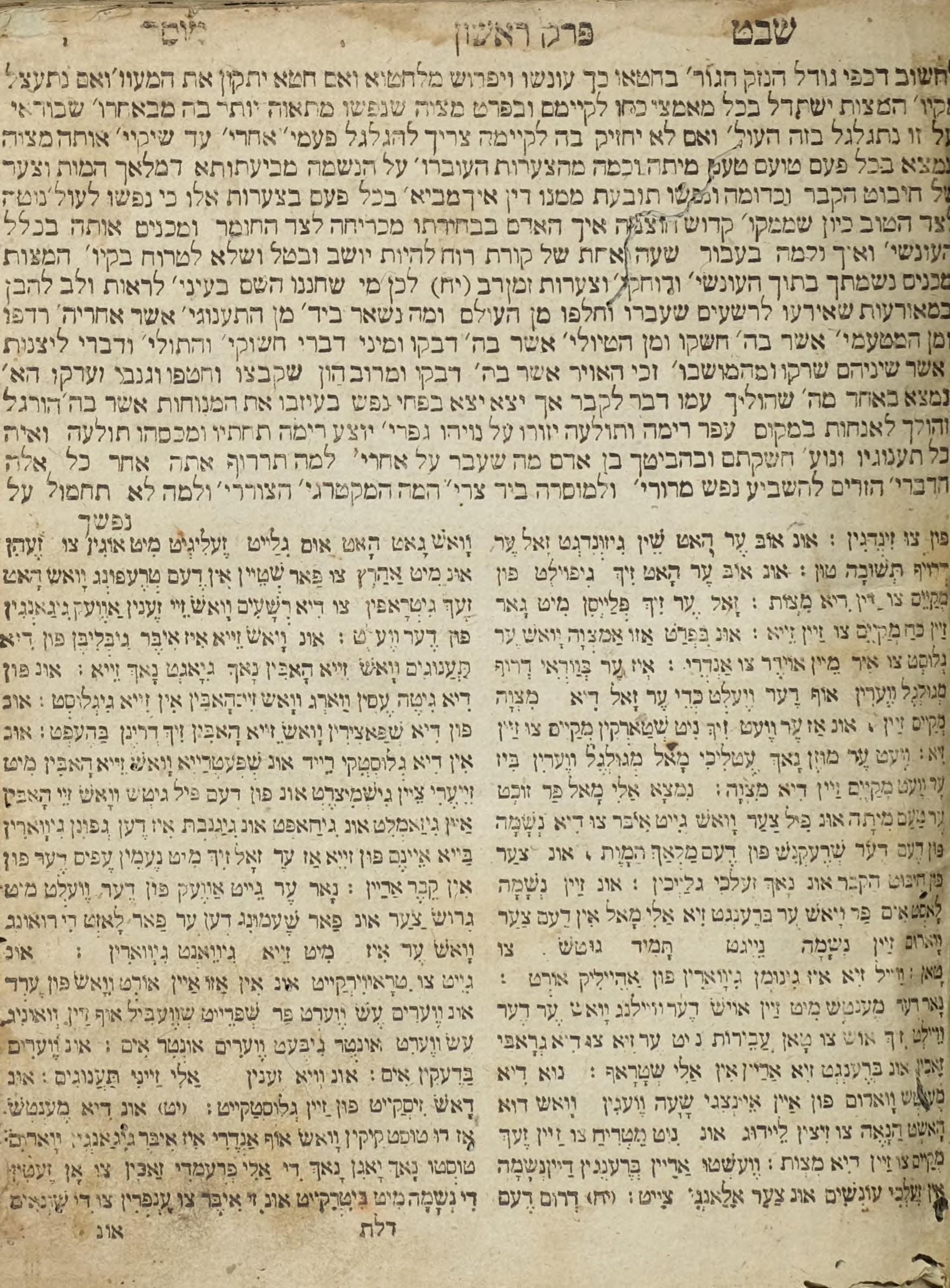} 
        \caption*{\UR}
        \label{fig:sub-u2}
    \end{subfigure}
    \hfill
    \begin{subfigure}{0.2\linewidth}
        \centering
        \includegraphics[height=24 mm]{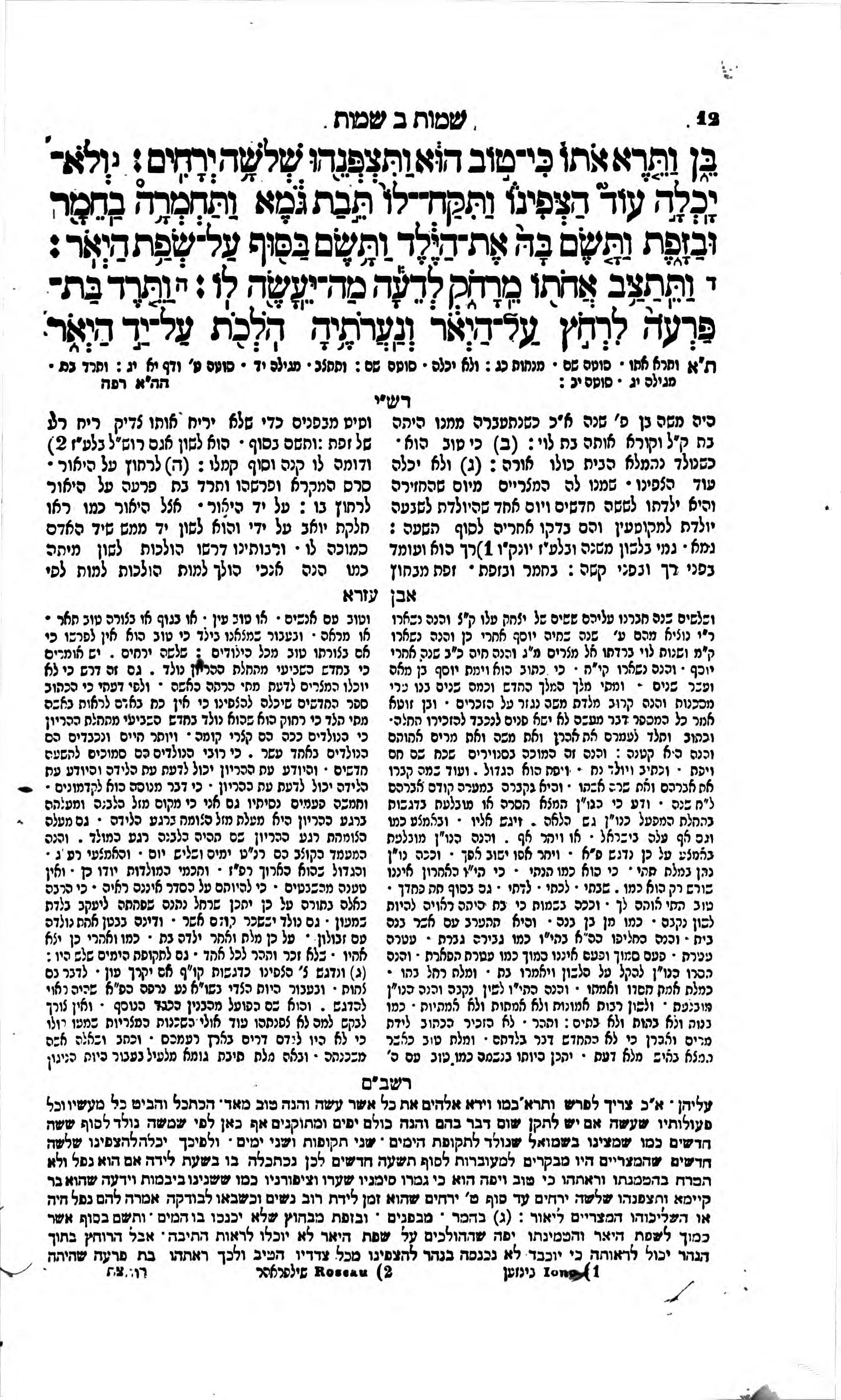} 
        \caption*{\OO}
        \label{fig:sub-o}
    \end{subfigure}
    \hfill
    \begin{subfigure}{0.2\linewidth}
        \centering
        \includegraphics[height=24 mm]{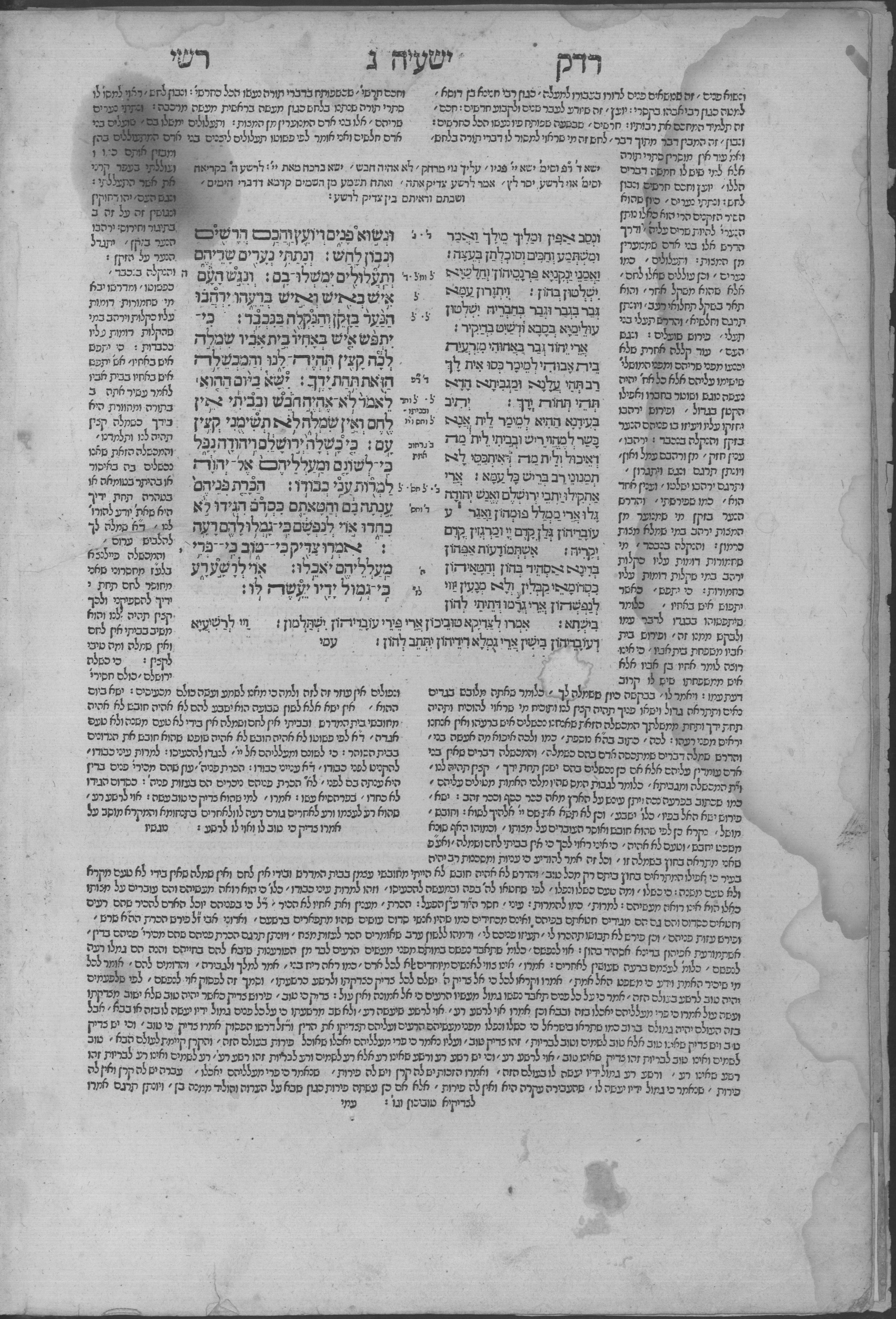} 
        \caption*{\YY}
        \label{fig:sub-y}
    \end{subfigure}
    \caption{Samples of complex layout classes.}
    \label{fig:complex_dataset}
\end{figure}

Two natural layouts for a page containing a main text along with a translation or commentary are two columns or top and bottom.
We refer to these layouts, as well as one region per page, as "simple."
But more complex options suggest themselves, especially when the main part is considerably shorter than the supplementary parts.
Rather than two vertical columns, the supplement may be wrapped around the main text in several ways: \LL-shaped and its mirror image (frequently on facing pages); \CC-shaped and its mirror image (already attested in the 8th century Codex Zacynthius, containing the Gospel of Luke with a running catena commentary); \OO-shaped with the main text in the center.
These arrangements culminate in the Glossa Ordinaria or "Talmudic" layout 
%\cite{Talmudic} 
(included in our catch-all \YY layout), with three (or more) main components: a central text, with one commentary on the outside (\LL or \CC) and another on the inside (\LL or \CC), with facing pages usually having a mirrored layout.
Also, instead of a full-width main text on the top of the page with upper and lower texts, the upper alone may be split into two columns in what we call the \UU format, or likewise, the lower text may be split in an inverted \UU format. 
When there are three components, top, bottom, and middle, one or more may be split into two columns.
This variety of layouts continued to be used when books with such texts were printed.
Fig.~\ref{fig:complex_dataset} presents illustrative examples of (Hebrew) books for each class.

Of special interest for manuscripts is the layout of the colophon page.
%(the scribe's note, often placed at the end of the manuscript). 
These are unique for every manuscript, but their classification, as well as that of the many other decorative formats (e.g.\@  \textit{tashqil} in Samaritan manuscripts or micrography in Hebrew ones), 
are beyond the scope of this paper.

Despite strong progress in document layout analysis, most benchmark datasets focus on layout analysis for English-language articles and similarly standardized document types. 
The recent advancements in deep learning-based layout analysis have been driven by large-scale annotated benchmarks like PubLayNet~\cite{publaynet} and DocLayNet~\cite{doclaynet2022}, which predominantly focus on English-language scientific articles or modern document types. 
For instance, in \cite{brazil},  document page classification is addressed using deep neural networks that leverage visual layout cues to distinguish broad structural categories.
However, these improvements do not easily apply to low-resource historical collections with complex layouts. Fully supervised methods need large amounts of annotated data and often fail to handle unseen layout structures. 
\rev{Digitizing such material faces many challenges, and supervised methods trained on available datasets struggle with the unseen, non-rectangular layouts of medieval manuscripts.}

% \begin{figure}[t]
%     \centering
%     \begin{subfigure}[b]{0.4\textwidth}
%         \centering
%         \includegraphics[width=0.5\linewidth]{figures/C_IE106037138_00025-OUT.png}
%         \caption{\CC type layout.}
%         \label{fig:sub1b}
%     \end{subfigure}\hfill
%     \begin{subfigure}[b]{0.4\textwidth}
%         \centering
%         \includegraphics[width=0.75\linewidth]{figures/U_IE112348603_00007-OUT.png}
%         \caption{\UU type layout.}
%         \label{fig:sub2b}
%     \end{subfigure}
%     \caption{A few samples with faulty segmentation by kraken/eScriptorium.}
%     \label{Fig2}
% \end{figure}

\begin{figure}[t]
    \centering
    \begin{subfigure}[b]{0.4\textwidth}
        \centering
        \includegraphics[width=0.4\linewidth]{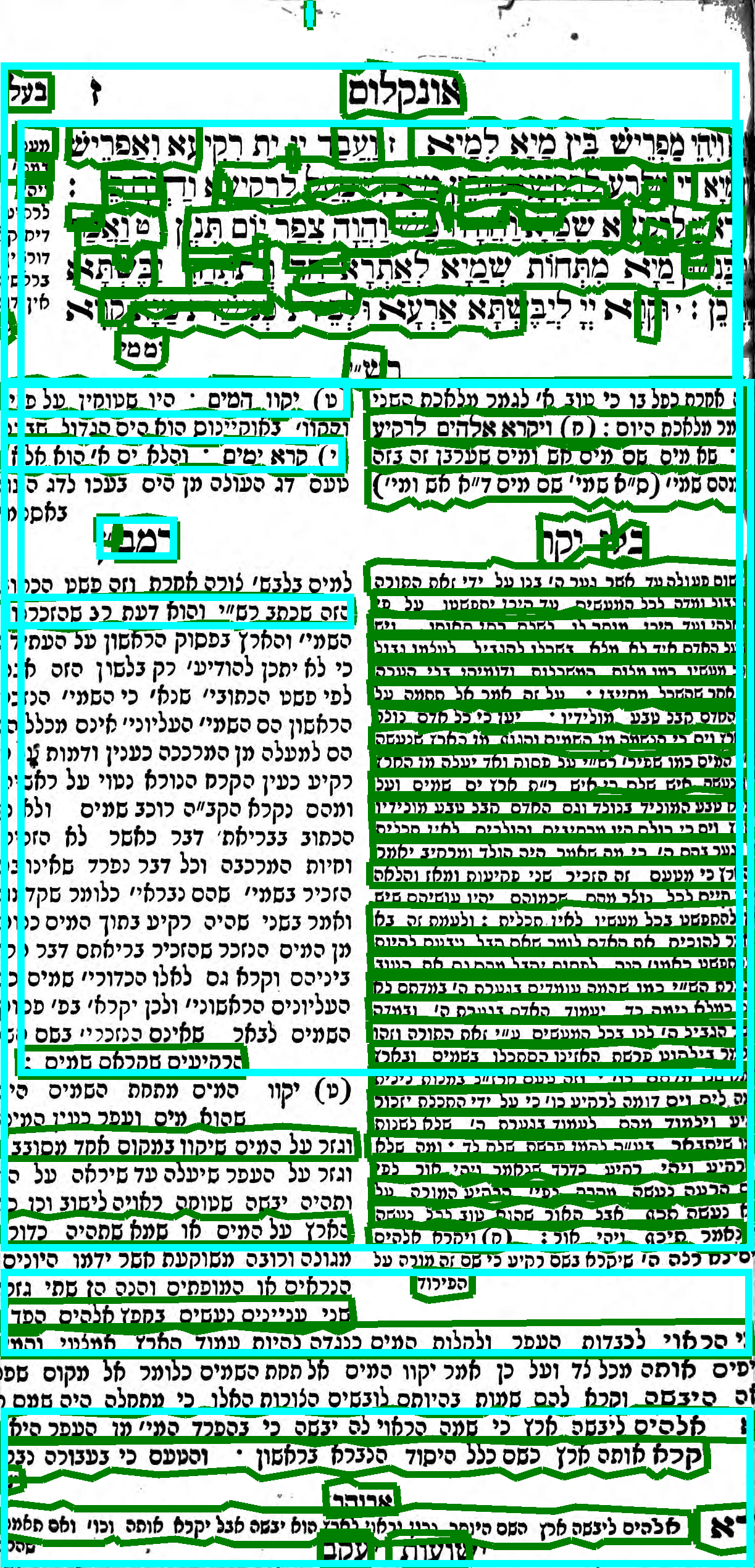}
    \end{subfigure}
    \begin{subfigure}[b]{0.4\textwidth}
        \centering
        \includegraphics[width=0.7\linewidth]{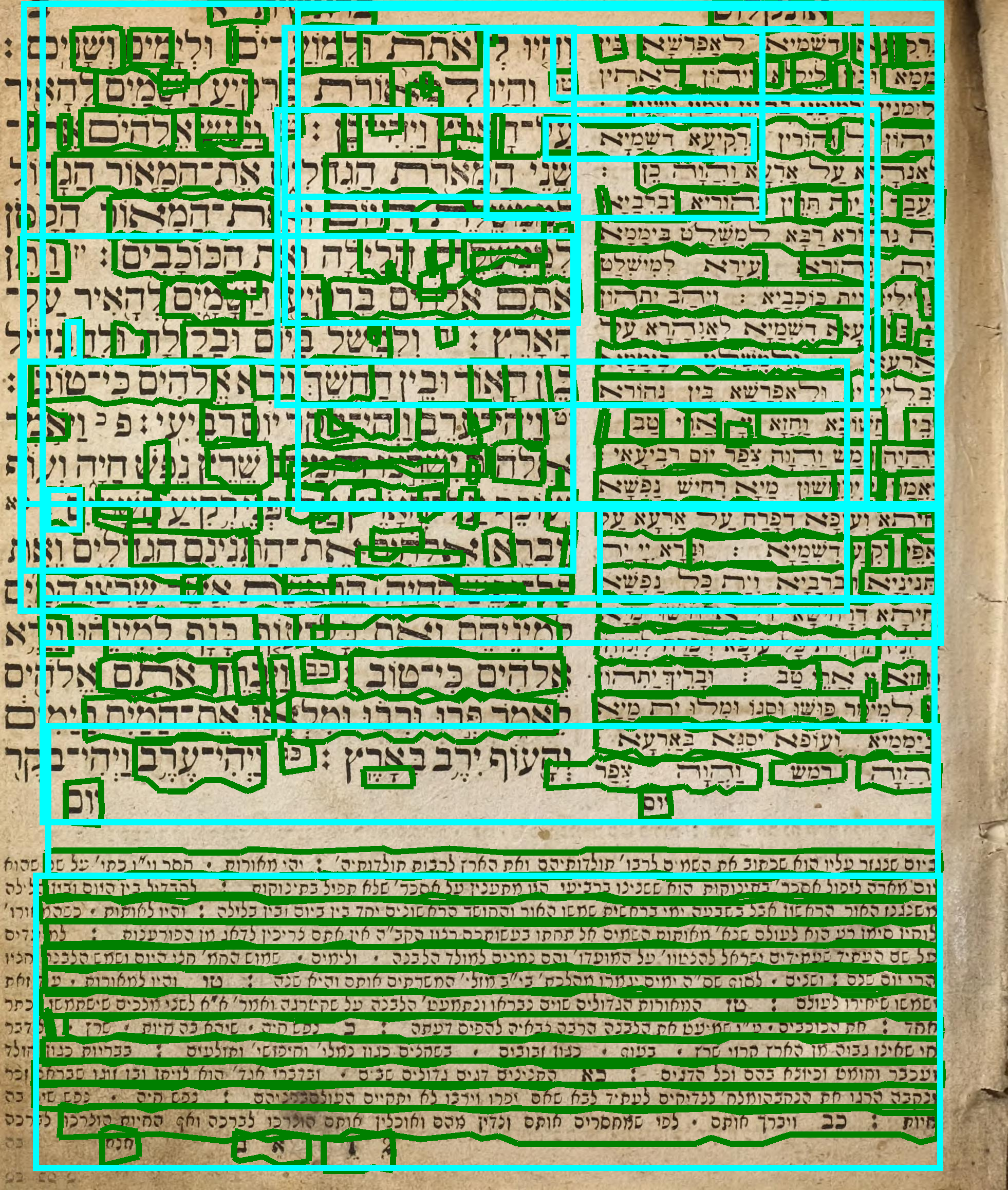}
    \end{subfigure}
    \caption{Complex layout samples with faulty segmentation.}
    %by kraken/eScriptorium.}
    \label{Fig2}
\end{figure}

It has been shown that pixel-based segmentation approaches often perform poorly when dealing with complex document layouts, especially when simple structural assumptions fail to capture the diverse organization of the page~\cite{m6doc}. Consequently, performance degradation is commonly observed in datasets containing intricate layout configurations. 
Even popular open-source text recognition environments like kraken %~\cite{Kiessling2019Kraken} 
and eScriptorium~\cite{kraken},
%\cite{Kiessling2021eScriptorium} 
which are designed for historical and non-Latin scripts, rely mostly  on pixel-level segmentation. 
This makes it hard for them to reliably separate meaningful regions in documents with complex layouts. As illustrated in Fig.~\ref{Fig2}, these systems often fail to distinguish between semantically distinct text blocks in complex \CC- or \LL-shaped configurations.
Thus, there is a critical need for robust classification models capable of learning from minimal samples without requiring massive-scale annotation efforts. 
These limitations motivate methods that can learn from tiny annotated datasets while focusing on stable geometric cues such as separator structure rather than overfitting to incidental textual content or style.

To address these challenges, we study complex-layout classification in a low-resource historical setting by curating a compact dataset and proposing a training strategy that emphasizes layout-defining separator geometry. 
The main contributions are as follows:
\begin{enumerate*}%[label=$\bullet$]
    \item[(1)]  We curate the CLC dataset, a specialized corpus of 155 high-resolution Hebrew pages manually annotated with eight distinct, semantically complex layout classes.

    \item[(2)]  We propose a \textit{Narrow Anisotropic Gaussian Augmentation} strategy that suppresses incidental textual details while preserving essential separator structures, compelling the model to learn global geometric arrangements rather than overfitting to font or content.

    \item[(3)]  \rev{We provide a systematic evaluation under severe data scarcity, including backbone comparison, augmentation ablation, external blind-corpus evaluation and a preliminary label-efficiency analysis.}
    
\end{enumerate*}

\begin{ignore}
    %The rest of the paper is organized as follows: 
Section 2 reviews relevant methods in layout analysis. Section 3 details the curation and characteristics of our  dataset and is followed by a description of our methods, including anisotropic masking. Section 5 reports on experimental results and comparative analysis, and Section 6 concludes with future directions.
\end{ignore}

%This research was conducted by our research team at the Tel Aviv University, as part of the MiDRASH ERC Synergy project for computational research of medieval Hebrew manuscripts \url{https://www.midrash.eu}. 
\section{Related Work}

\paragraph{Layout analysis benchmark datasets.}
Layout analysis has benefited substantially from the release of large-scale annotated benchmarks. 
PubLayNet~\cite{publaynet} is a widely used public dataset built primarily from modern scientific articles, enabling strong supervised training for document segmentation and structural understanding. 
DocLayNet~\cite{doclaynet2022} expands this direction by covering more diverse and visually complex layouts drawn from a variety of public document sources. 
The PRImA~\cite{prima} layout analysis dataset is comprised of heterogeneous document types (e.g.\@ magazines and technical articles), comprising 305 pages grouped into six categories, namely, Financial Reports, Manuals, Scientific Articles, Laws \& Regulations, Patents, and Government Tenders;
 the majority of documents are in Latin scripts. 
More recently, M\textsuperscript{6}Doc~\cite{m6doc} explicitly targets diversity in both layout and language, including multi-layout settings (rectangular, Manhattan, non-Manhattan, and multi-column Manhattan) and \rev{multilingual} content (Chinese and English). Collectively, these datasets have shaped the dominant evaluation protocols and model designs for layout analysis under modern, well-resourced conditions.

\paragraph{Historical layout datasets.} %Across Languages.}
Several datasets have been compiled to address historical documents with unique challenges due to noise, degradation, and layout complexity.  NewsEye~\cite{news} comprises Austrian newspaper pages from the 19th and early 20th century with carefully corrected text, including 148 training pages and 13 validation pages, with ground truth created using Transkribus by the Austrian National Library. DIVA-HisDB~\cite{diva} includes 150 pages from three medieval Latin manuscripts from the Carolingian period. A Brazilian historical newspaper dataset contains 140 page images sourced from the Brazilian National Digital Library (BND)~\cite{brazil}. 
For non-Latin scripts, SCUT-CAB~\cite{scut} addresses the complex layouts of ancient Chinese books with 4000 manually annotated images, while the HJDataset~\cite{hjd} provides a large-scale resource for historical Japanese documents, including hierarchical structures and reading-order information. 
These datasets highlight the growing recognition that historical layout analysis requires specialized resources; they also underscore how fragmented the landscape remains across scripts and traditions.
U-DIADS-Bib~\cite{udiads} introduced a few‑shot benchmark for ancient manuscript layout segmentation. The ICDAR 2025 FEST competition~\cite{fest} targeted few‑shot text‑line segmentation of handwritten documents. 
Our focus is on page‑level separator‑driven layout classification as an upstream routing step before segmentation and OCR.

\paragraph{Page-level classification.}
Approaches to page-level layout classification have evolved from feature engineering to deep learning. Early work often relied on manually engineered features. A study on page layout classification for legal documents categorized pages into structural types such as single column, marginal single column, and two column formats. The results showed that feature-engineered methods (e.g.\@ SVM and random forest) leveraging geometric layout cues achieved accuracies around 94\%, emphasizing the value of explicit structural features~\cite{feature}. With the advent of deep learning, convolutional neural networks (CNNs) became the dominant strategy. 
A prior study demonstrated that CNNs could effectively discriminate between diverse categories such as letters, memos, invoices, etc., by analyzing whole-page visual cues~\cite{prima}. 
Additionally, frameworks for graphical object detection~\cite{graphical} have leveraged transfer learning and domain adaptation to identify graphical elements like tables and figures, mitigating the scarcity of annotated training data.

\paragraph{Hebrew layout resources.} 
In the specific context of Hebrew document layout analysis, resources are notably scarce. 
The primary existing benchmark is NetLay~\cite{gogawale2024netlay}, which contains approximately 1,300 pages from printed Hebrew (or Hebrew-alphabet) books. 
NetLay utilizes a multi-label encoding approach with VGG16 to classify pages into four categories, namely no-text, single-column, double-column, and complex.
However, the label space remains relatively coarse with respect to separator-defined, non-rectangular historical layouts (e.g.\@ \CC- or \LL-shaped configurations), which motivates finer-grained datasets and modeling assumptions that explicitly target separator geometry.
Motivated by these gaps, our work focuses on separator-defined complex layout classification for low-resource historical Hebrew pages, complementing existing Hebrew resources (e.g.\@ NetLay) with a finer-grained label space tailored to non-rectangular and multi-region structures.  Our goal is to provide a targeted, publicly usable dataset and baseline that isolates the layout-classification problem under severe annotation constraints and emphasizes global geometric structure.

\section{Complex Layout Classification (CLC) Dataset}

We curate a compact, task-specific corpus to address complex layout classification with the high-level goal of improving downstream text transcription systems by routing pages to layout-aware pipelines. The dataset comprises 155 high-resolution document images sourced from the National Library of Israel’s digital collections and printed books from various periods. To evaluate the efficacy of our model in a low-resource setting, we work with a small but representative dataset, a common constraint in computational humanities where large-scale annotated data are scarce.

The images are manually categorized into our  eight classes. 
The complexity of these layouts is not merely structural but also semantic; pages often contain separate but related works by different authors, thereby necessitating models that can capture complex, multi-block hierarchical relationships. 
They use two main script types: square (block letters) and "Rashi" (curvilinear).
To support robust learning from small data, all images are retained at their native scan resolution. 
Table~\ref{tab:dataset} summarizes 
the distribution across the eight layout classes, the script types present 
in each, and the annotation source. %\todo{@Daria can u help me complete this?}
Fragments in non-Hebrew characters are a common phenomenon in complex layouts, as such texts typically include commentaries or translations into local vernaculars or appear in scholarly editions.

Ground truth annotations were established through a rigorous protocol. To maximize structural diversity, candidate pages were restricted to a single representative page per printed book. Initial categorizations were performed by trained annotators, and all labels were subsequently reviewed and verified by an expert paleographer, resolving the disagreements to produce a single ground-truth label per image.
The complete dataset, including the images, book identifiers, and the ground truth,  is available at \url{https://github.com/TAU-CH/midrash_clc}.

\begin{table}[t]
\centering
\caption{CLC dataset statistics. Sq = square (block letters); 
Ra = Rashi (curvilinear). Pages sourced from the National Library of 
Israel digital collections.}
\label{tab:dataset}
\setlength{\tabcolsep}{5pt}
\begin{tabular}{lcrc}
\toprule\small
\textbf{Class} & \textbf{Count} & \textbf{\% Total} & 
\textbf{Script Types} \\
\midrule
\CC  & 21 & 13.5 & Sq, Ra  \\
\CR  & 19 & 12.3 & Sq, Ra  \\
\LL  & 28 & 18.1 & Sq, Ra  \\
\LR  & 22 & 14.2 & Sq, Ra  \\
\OO  & 22 & 14.2 & Sq, Ra   \\
\UU  & 16 & 10.3 & Sq, Ra  \\
\UR  & 13 &  8.4 & Sq, Ra    \\
\YY  & 14 &  9.0 & Sq, Ra  \\
\midrule
\textbf{Total} & \textbf{155} & \textbf{100.0} & ---  \\
\bottomrule
\end{tabular}
\end{table}

%\todo{\url{https://drive.google.com/drive/folders/1cLtAf9tPU_BObwJrWPYQjyNs7U45CQtP} The Y folder contains a few examples of Q as well.
%\url{https://drive.google.com/drive/folders/1cLtAf9tPU_BObwJrWPYQjyNs7U45CQtP} is this Y or C}

\section{Method}

We address the task of complex-layout classification, where document images are to be assigned to one of eight predefined layout types. Let $\mathcal{C}$ denote the set of layout classes,
$
\mathcal{C} = \{\CC, \CR, \LL, \LR, \UU, \UR, \OO, \YY\}.
$
These classes correspond to page layouts containing text regions shaped as \CC, horizontally reflected \CC, \LL, horizontally reflected \LL, \UU, vertically reflected \UU, or \OO, with irregular region forms grouped under the \YY class.

The primary challenge in classifying document images into these eight classes is the severe scarcity of annotated data, with only a handful of samples available per class. To address this limitation, we propose a CNN-based framework that employs strong augmentation strategies to improve generalization.

In layout classification, the defining feature of each class lies in its separator regions, which partition the page into characteristic geometric layouts defined in $\mathcal{C}$. In contrast, the textual content, including font type, size, or semantic information does not determine the layout category. Therefore, any augmentation procedure that preserves separator regions encourages the model to focus on the global arrangement of separators rather than incidental textual details.

In particular, we adopt two complementary families of augmentations: (1) Narrow anisotropic Gaussian based masking, which applies multi-oriented anisotropic Gaussian filters with a narrow standard deviation 
%of $\sigma_x = 1$ 
to suppress text regions while preserving separator regions; and (2) horizontal and vertical reflections, which enrich the diversity of training samples while ensuring label consistency in asymmetric categories. Together, these augmentations significantly expand the effective training distribution and enable robust CNN learning despite the limited availability of annotated data.

\begin{figure}[t]
\centering
\includegraphics[width=0.25\linewidth]{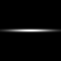}
\qquad\qquad
\includegraphics[width=0.25\linewidth]{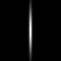}
\caption{Narrow anisotropic Gaussians used for masking ($\sigma_x = 1$; for visualization, $\sigma_y = 15$ is shown). 
Left: horizontal ($\theta=0$). 
Right: vertical ($\theta={\pi}/{2}$).}
\label{fig:gaussian-filters}
\end{figure}

\paragraph{Narrow anisotropic Gaussian based masking.}
We augment the  document images using a multi-oriented anisotropic Gaussian filter that preserves separator regions. For orientation $\theta$ and standard deviations $\sigma_x,\sigma_y$, the kernel is
\[
G_{\sigma_x,\sigma_y,\theta}(x,y) \;=\; 
\frac{1}{2\pi \sigma_x \sigma_y}
\exp\!\left(
-\frac{(x\cos\theta + y\sin\theta)^2}{2\sigma_x^2}
-\frac{(-x\sin\theta + y\cos\theta)^2}{2\sigma_y^2}
\right)
\]
Given a document image $I(x,y)$, where $(x,y)$ are pixel coordinates, containing low-contrast text on a high-contrast background, we fix $\sigma_x = 1$ and filter $I(x,y)$ with $G_{1,\sigma_y,\theta}$ at orientations $\theta \in \{0, {\pi}/{2}\}$ (Fig.~\ref{fig:gaussian-filters}).
For each pixel, the maximum response across the two orientations is computed to form the response map $R_{\sigma_y}(x,y)$. The directional response and the subsequent binary mask are defined as:
\smash{$   R_{\sigma_y}(x,y) = \max_{\theta \in \{0, {\pi}/{2}\}} (I * G_{1,\sigma_y,\theta})(x,y)$}; \smash{$
M_{\sigma_y}(x,y) = \mathbf{1}[\tilde{R}_{\sigma_y}(x,y) \ge t]
$},
where \smash{$\tilde{R}_{\sigma_y}$} denotes the min-max normalized response map scaled to $[0,1]$, and $\mathbf{1}[\cdot]$ is the indicator function. We then form the masked augmentation by preserving pixels in separator regions and suppressing the remaining content:
$I'(x,y) = 255 \cdot M_{\sigma_y}(x,y) + \big(1-M_{\sigma_y}(x,y)\big)\,\big(255-I(x,y)\big)$.
Pixels outside the separator mask are inverted to suppress fine-grained text texture while retaining  global page structure and high-contrast separator geometry.

\begin{figure}[t]
    \centering
    \includegraphics[width=0.6\textwidth]{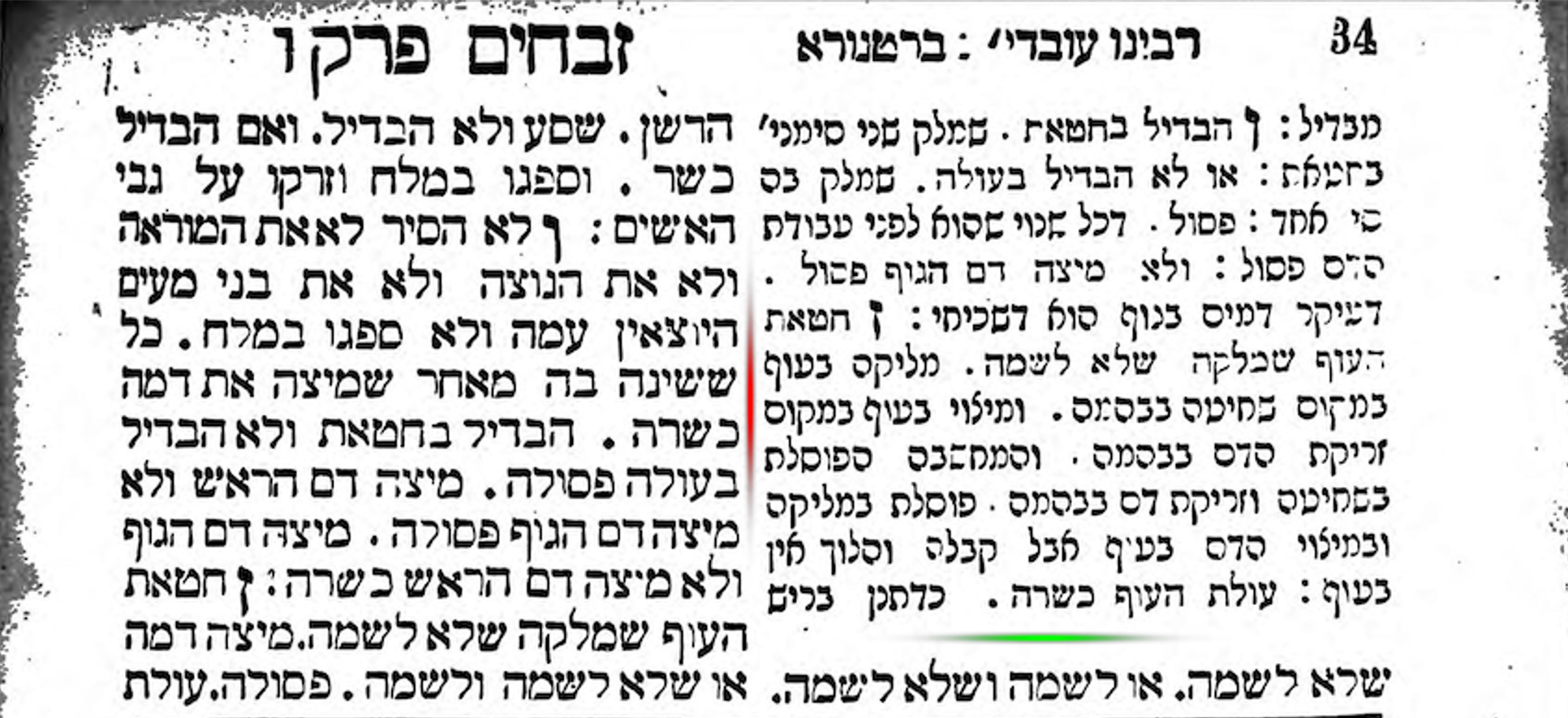}
    \caption{Illustration of separator region preservation in a document with an \LL-shaped layout. On vertical separator regions, no matter how narrow they are, a narrow vertical Gaussian (red) will lead to a high response. Similarly, on horizontal separator regions, no matter how narrow they are, a narrow horizontal Gaussian (green) will lead to a high response.}
    \label{fig:nag_separator}
\end{figure}

The use of $\sigma_x = 1$ together with multi-orientation filtering ensures that separator region pixels consistently yield high responses. 
In our case, we set $\theta \in \{0, {\pi}/{2}\}$, since separator regions in the dataset are oriented only vertically or horizontally. At vertical separator regions, the maximum response is produced by the vertically oriented narrow Gaussian, whereas at horizontal separator regions it is produced by the horizontally oriented narrow Gaussian. Because the filters are narrow, their placement at separator regions near text region boundaries still yields high responses without blurring, despite the presence of neighboring low-contrast pixels within the text regions (Fig.~\ref{fig:nag_separator}).

\paragraph{Horizontal and vertical reflections.}
To further enrich the training distribution, we apply horizontal and vertical reflections, which preserve separator regions while modifying the overall page structure. In contrast, rotations and shearings are deliberately excluded, as they alter separator orientations and thereby disrupt the structural integrity of the layout. Translations are avoided, since they may shift crucial separator regions outside the field of view, invalidating the class-defining structure. And scalings are omitted because all input images are rescaled to a fixed size prior to training, rendering additional scaling redundant.

Reflections do require relabeling; however, the label set is closed under reflection, ensuring that the new labels remain within the 
%same predefined class set 
$\mathcal{C}$.
%as depicted in Table~\ref{tab:reflection-mapping}.
%Specifically, when a document image belonging to class $c \in \mathcal{C}$ is reflected through an axis $a$, if $c$ is asymmetric with respect to $a$, then the reflected image is relabeled with the corresponding reflection label $c^{a} \in \mathcal{C}$. Conversely, if $c$ is symmetric with respect to $a$, the label remains unchanged, i.e., $c^{a} = c$. The complete set of label transformations under vertical and horizontal reflections is summarized in Table~\ref{tab:reflection-mapping}.
%Specifically, 
Horizontal reflection exchanges $\CC\leftrightarrow\CR$
and $\LL\leftrightarrow\LR$.
Vertical reflection exchanges $\UU\leftrightarrow\UR$
and turns \LL and \LR into \YY.
Two classes, \OO and \YY, are (usually) retained under reflection.

\begin{ignore}[t]
\caption{Reflection induced label transformations for all layout classes in $\mathcal{C}$. Classes change label under vertical or horizontal reflection if they are asymmetric with respect to the corresponding axis.}
\label{tab:reflection-mapping}
\centering
\begin{tabular}{ccc}
\hline
\textbf{Original} & \textbf{Horizontal} & \textbf{Vertical} \\
\textbf{class} & \textbf{reflection} & \textbf{reflection} \\
\hline
\CC & \CR & \CC \\
\CR & \CC & \CR \\
\LL & \LR & \YY \\
\LR & \LL & \YY \\
\hline
\end{tabular}\qquad
\begin{tabular}{ccc}
\hline
\textbf{Original} & \textbf{Horizontal} & \textbf{Vertical} \\
\textbf{class} & \textbf{reflection} & \textbf{reflection} \\
\hline
\UU & \UU & \UR \\
\UR & \UR & \UU \\
\OO & \OO & \OO \\
\YY & \YY & \YY \\
\hline
\end{tabular}
\end{ignore}

\begin{figure}[t]
    \centering
    \includegraphics[width=0.7\textwidth]{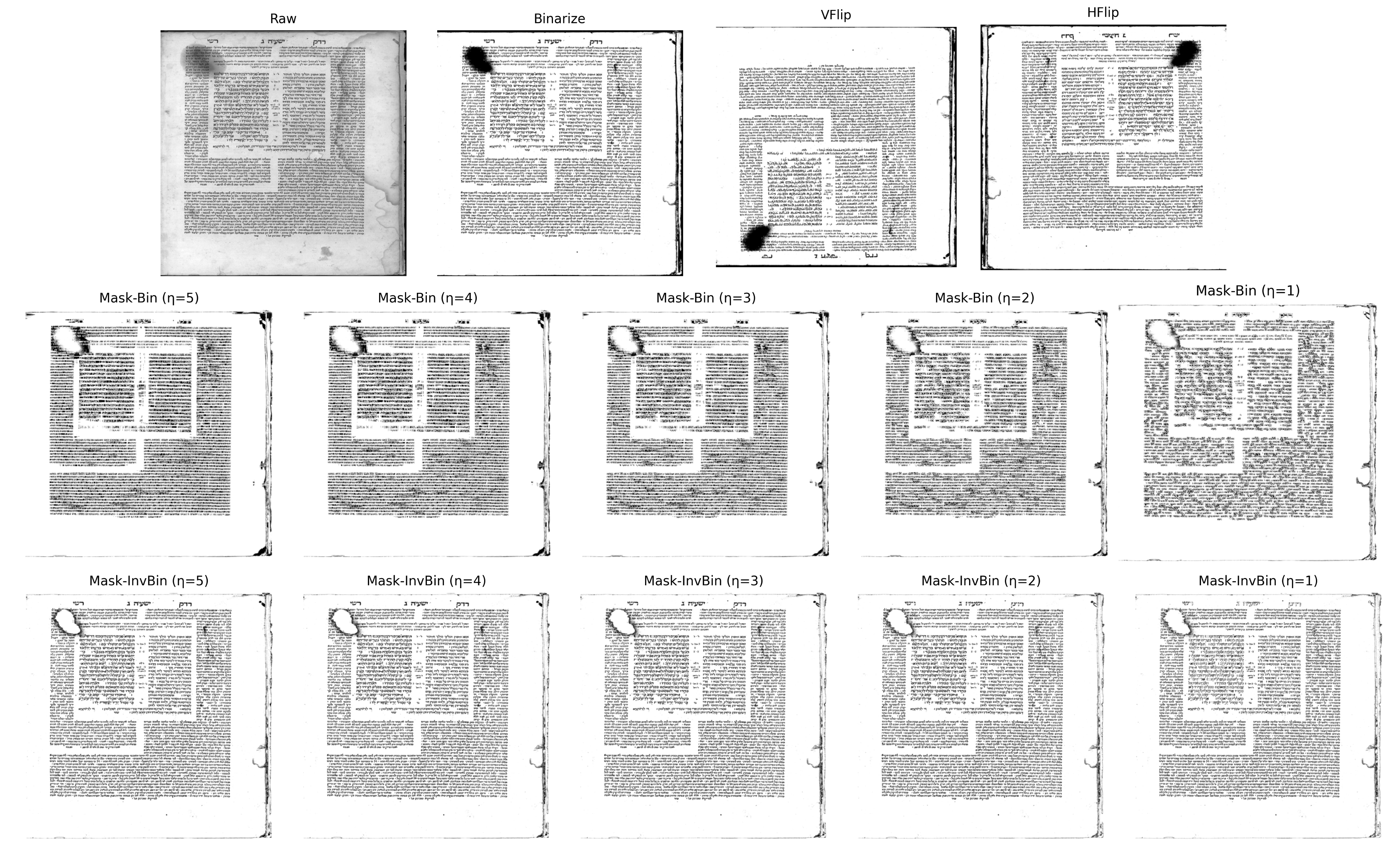}
    \caption{Comprehensive view of the augmentation suite. A single training image is expanded into eleven distinct intensity representations (binary, masked, and inverse-masked), alongside their valid spatial reflections.}
    \label{fig:aug}
\end{figure}

\paragraph{Augmentations.}
To overcome severe data scarcity, we implement a combinatorial 
but task-constrained augmentation suite that preserves class-defining 
separator topology, in contrast to generic appearance perturbations 
which perturb arbitrary image regions 
without structural awareness.
For each training image, we first binarize, while  also extracting an inverse-binarized variant to minimize polarity sensitivity. We then generate masking-based variants by applying our narrow anisotropic Gaussian operator across five strength settings, $\eta \in \{1,2,3,4,5\}$. This gives both binarized-masked and inverse-binarized-masked modalities, resulting in a transformed pool of $|\mathcal{T}| = 11$ distinct intensity representations per image (1 plain binary, 5 masked binary, and 5 inverse-masked binary) as illustrated in Fig.~\ref{fig:aug}. Subsequently, each of the 11 modalities is subjected to a spatial reflection protocol using deterministic label mappings. Horizontal reflection is universally applied, exchanging asymmetric pairs,
%(e.g.\@ $\CC \leftrightarrow \CR$, $\LL \leftrightarrow \LR$) 
while preserving symmetric classes. Vertical reflection is applied exclusively to the subset of classes where structural semantics remain well-defined (\UU, \UR, \OO, \CC, \CR, \YY). By enumerating all valid spatial and intensity combinations, this pipeline effectively acts as a dataset multiplier, yielding 22 to 33 training instances per original image depending on class validity under vertical reflection. Finally, to mitigate the residual class imbalance within this expanded set, we employ weighted random sampling during optimization.

\paragraph{ConvNeXt-based classifier.}
ConvNeXt~\cite{convnext} is a modernized Convolutional Network which retains inductive biases that matter for layout: locality, translation equivariance, and hierarchical composition---properties well aligned with separator-driven global geometry under scarce labels. A ConvNeXt block replaces classic ResNet bottlenecks with: (i) large-kernel depthwise convolution for long-range spatial aggregation, and (ii) pointwise “MLP-style” mixing for channel interactions, while using LayerNorm/GELU for stable optimization.

Let \smash{$x \in \mathbb{R}^{H \times W \times C}$} be an input feature map. A depthwise convolution with kernel $K$ operates per-channel:
\begin{equation*}
\mathrm{DWConv}(x)_{i,j,c} = \sum\nolimits_{u,v \in K} w_{u,v,c}\; x_{i+u,\,j+v,\,c}.
\label{eq:dwconv}
\end{equation*}
Channel mixing is then performed by pointwise projections (equivalent to $1{\times}1$ convolutions):
For a channel vector $\mathbf{z}_{i,j} \in \mathbb{R}^C$ at spatial location $(i,j)$, this operates as a linear transformation
$\mathrm{PW}(\mathbf{z})_{i,j} = \mathbf{W} \mathbf{z}_{i,j}$.
A simplified ConvNeXt block can be written as:
\begin{equation*}
y = x + \mathcal{F}(x), \qquad
\mathcal{F}(x) = \mathrm{PW}_{2}\!\left(\sigma\!\left(\mathrm{LN}\!\left(\mathrm{PW}_{1}\!\left(\mathrm{DWConv}(x)\right)\right)\right)\right),
\label{eq:convnext_block}
\end{equation*}
where $\mathrm{LN}$ is LayerNorm, $\sigma$ is the GELU activation, $\mathrm{PW}_{1}$ is an expansion layer that increases the channel dimension by a factor of 4, and $\mathrm{PW}_{2}$ is a projection layer that reduces it back to the original dimension.

For classification, the network outputs logits $g(x)\in\mathbb{R}^{8}$, and is trained with standard cross-entropy loss.

% :\todo{omit equation}
% \begin{equation*}
% \mathcal{L} = -\frac{1}{N}\sum\nolimits_{n=1}^{N} \log
% \frac{\exp\big(g(x_n)_{y_n}\big)}{\sum_{k=1}^{8} \exp\big(g(x_n)_{k}\big)}.
% \label{eq:ce}
% \end{equation*}

We fine-tune an ImageNet-pretrained ConvNeXt-Tiny, adapting the stem to 1-channel inputs by averaging RGB pretrained weights, and replacing the final classifier for 8 classes. We use \emph{staged fine-tuning}: we train the classifier head first and then unfreeze the backbone with a smaller learning rate, which reduces overfitting early and improves stability under scarce supervision.
We first establish a strong \emph{supervised baseline} by fine-tuning standard image backbones for 8-way page-level layout classification (\CC, \CR, \LL, \LR, \UU, \UR, \OO, \YY).

Input images were resized to 224$\times$224 pixels and binarized using Otsu's thresholding prior to ingestion. We utilized a stratified data split, allocating 60\% to training, 20\% to validation, and 20\% to testing. To prevent destabilizing the pretrained weights, we employed a two-stage fine-tuning strategy for a maximum of 100 epochs using a batch size of 16. For the first 10 epochs, we froze all non-classifier parameters and trained the head using the AdamW optimizer (initial learning rate $10^{-3}$, weight decay $10^{-4}$) with a cosine annealing schedule ($T_{\max} = 70$, $\mathrm{lr}_{\min} = 10^{-5}$). Subsequently, the entire network was unfrozen, and AdamW was reinitialized with a lower learning rate of $10^{-4}$ (weight decay $10^{-4}$) and a new cosine schedule ($T_{\max} = 90$, $\mathrm{lr}_{\min} = 10^{-6}$). Training was regulated by early stopping with a patience of 10 epochs based on validation loss.

\paragraph{Active learning protocol.} % for Label-Efficient Layout Annotation.}
\label{subsec:method_al}
We also study an active learning protocol to evaluate label efficiency in our low-resource setting, which reflects historical document analysis, where structurally complex layouts often require domain-expert annotation and labels are expensive to obtain. Therefore, even modest reductions in labeling effort are practically useful.

Let $\mathcal{D}_{\text{train}}$ be the training split. At round $t$, we maintain a labeled set $\mathcal{L}_t \subseteq \mathcal{D}_{\text{train}}$ and an unlabeled pool $\mathcal{U}_t = \mathcal{D}_{\text{train}}\setminus \mathcal{L}_t$. We initialize with $\rho_0 \approx 0.2$ and add $\Delta\rho \approx 0.2$ each round until the pool is exhausted. At each round, we train ConvNeXt-Tiny with our \textbf{Full} pipeline (Binary + Reflections + Anisotropic Masking) on the labeled set, then select a batch $\mathcal{S}_t \subset \mathcal{U}_t$ of size $B$ according to an acquisition function $a_t(x)$: $ \mathcal{S}_t = \argmax_{S\subseteq \mathcal{U}_t,\,|S|=B}\, \sum\nolimits_{x\in S} a_t(x), $ followed by $\mathcal{L}_{t+1}\leftarrow \mathcal{L}_t\cup \mathcal{S}_t$ and $\mathcal{U}_{t+1}\leftarrow \mathcal{U}_t\setminus \mathcal{S}_t$. We compare three simple acquisition rules: (i) \textbf{Random} ($a_t(x)$ uniform); (ii) \textbf{Entropy}, using predictive entropy $a_t(x) = -\sum_{c=1}^{8} p(c\mid x)\,\log\!\big(p(c\mid x)+\epsilon\big)$; and (iii) \textbf{Margin}, using the negative gap between the two highest class probabilities, $a_t(x) = -\big(p_{(1)}(x) - p_{(2)}(x)\big)$, where $p_{(1)}(x)$ and $p_{(2)}(x)$ denote the largest and second-largest entries of $p(\cdot\mid x)$. For the uncertainty rules we select the $B$ samples with the highest $a_t(x)$. 

\paragraph{Learning curve summary.} 
Across labeled budgets $\rho_t = |\mathcal{L}_t|/|\mathcal{D}_{\text{train}}|$, we evaluate accuracy and macro-$F_1$ on a fixed held-out split and summarize label-efficiency using the area under the learning curve (AULC).

\begin{ignore}

\section{Experiments}\todo{where is this section?}

\todo{Add here an Evaluation subsection. Define the metrics used for evaluation.}
\todo{Include the ablations here}
\todo{Include the active learning here}
\end{ignore}
\section{Results}

\paragraph{Backbone architecture comparison.}
To determine the optimal architecture for complex layout classification under severe data constraints, we evaluated a diverse set of deep learning models: classical CNNs (VGG16, ResNet-18), a highly optimized mobile architecture (EfficientNet-B0), a pure vision transformer (ViT-B/16), and a modernized CNN (ConvNeXt-Tiny). All architectures were trained using our proposed Full layout-preserving pipeline (Binary+Flips+Masking) to ensure a fair comparison. Because the test set is small (31 pages), we report macro-averaged precision, recall, and $F_1$-score alongside accuracy to account for class imbalance.

\begin{figure}[t]
    \centering
    \includegraphics[width=\textwidth]{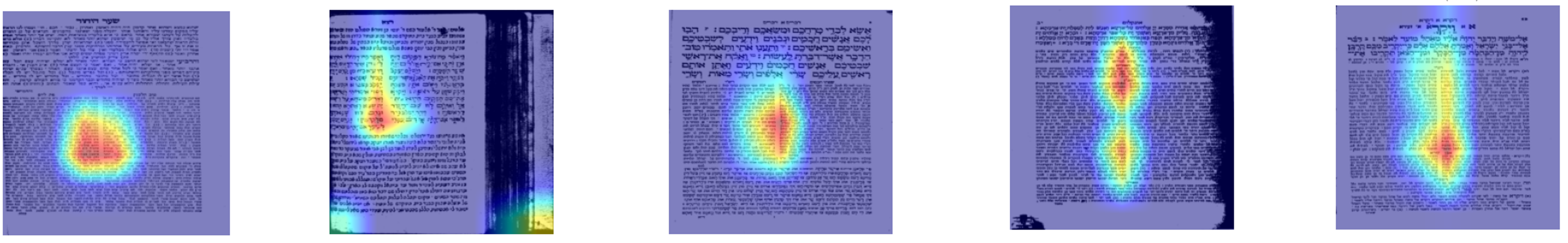}
    \caption{GradCAM visualizations of the ConvNeXt-Tiny classifier. Warmer colors indicate regions with higher contribution to the predicted layout class.}
    \label{fig:gradcam}
\end{figure}

Compared to vision transformers, which typically require extensive supervision to learn strong spatial priors, ConvNeXt relies less on learning global structure from scratch. 
The large receptive field from depthwise large-kernel convolutions allows the network to capture the global geometric arrangement of separator regions (e.g.\@ distinguishing a \UU layout from an \LL layout), and  parameter-efficient depthwise operations reduce overfitting risk and improve  efficiency. This dynamic is visually confirmed in Fig.~\ref{fig:gradcam}, where the network's attention is predominantly localized along separator regions, block boundaries, and inter-column gaps rather than dense textual content. Such qualitative evidence demonstrates that our proposed preprocessing and augmentations successfully compel the classifier to anchor its predictions on global separator geometry.

% \begin{table}[t]
% \centering
% \caption{Baseline backbone comparison on CLC (test set: 31 pages), reporting accuracy and macro-averaged precision, recall, and $F_1$.}
% \label{tab:backbone_baselines}
% %\setlength{\tabcolsep}{6pt}
% %\renewcommand{\arraystretch}{1.15}
% \begin{tabular}{lcccc}
% \hline
% \textbf{Model} & \textbf{Params} & \textbf{Acc.} & \textbf{Prec. (Macro)} & \textbf{Rec. (Macro)} & \textbf{$F_1$ (Macro)} \\
% \hline
% \textbf{ConvNeXt-Tiny} & 28.6M & \textbf{0.90} & \textbf{0.93} & \textbf{0.88} & \textbf{0.88} \\
% EfficientNet-B0   & 5.3M    & 0.45 & 0.39 & 0.44 & 0.41 \\
% VGG16   & 138.4M   & 0.45 & 0.39 & 0.44 & 0.41 \\
% ResNet-18  & 11.7M & 0.23 & 0.16 & 0.20 & 0.17 \\
% ViT-B/16  & 86.6M & 0.23 & 0.19 & 0.20 & 0.15 
% \end{tabular}
% \end{table}

\begin{table}[t]
\centering
\caption{Left: Backbone comparison under our full layout-preserving pipeline on the CLC dataset. Right: NetLay-adapted multi-label baseline models \emph{without} layout-preserving augmentations. 
Test set: 31 pages. Best in bold.}
\label{tab:backbone_baselines}
\begin{minipage}{0.54\linewidth}
\centering
\begin{tabular}{lrcccc|}
\toprule
\textbf{Model} & \textbf{Params} & \textbf{Acc.} & \textbf{P} & \textbf{R} & \textbf{$F_1$} \\
\midrule
\textbf{ConvNeXt-Tiny} & 28.6M & \textbf{0.90} & \textbf{0.93} & \textbf{0.88} & \textbf{0.88} \\
EfficientNet-B0        &  5.3M & 0.45 & 0.39 & 0.44 & 0.41 \\
VGG16                  & 138M  & 0.45 & 0.39 & 0.44 & 0.41 \\
ResNet-18              & 11.7M & 0.23 & 0.16 & 0.20 & 0.17 \\
ViT-B/16               & 86.6M & 0.23 & 0.19 & 0.20 & 0.15 \\
\bottomrule
\end{tabular}
\end{minipage}
\hfill
\begin{minipage}{0.42\linewidth}
\centering
\begin{tabular}{lcccc}
\toprule
\textbf{NetLay} & \textbf{Acc.} & \textbf{P} & \textbf{R} & \textbf{$F_1$} \\
\midrule
EfficientNet-\\
V2-S & \textbf{0.52} & \textbf{0.60} & \textbf{0.48} & \textbf{0.49}   \\
VGG16             & 0.45 & 0.42 & 0.42 & 0.40 \\
ViT-B/16          & 0.23 & 0.30 & 0.19 & 0.17 \\
\bottomrule\\
\end{tabular}
\end{minipage}
\end{table}

\begin{table}[t]
\centering
\caption{External blind evaluation on the test
%[Midrash] 
corpus (20 samples per class).}
\label{tab:external_blind}
\begin{tabular}{cccc|cccc}
\toprule
Class & Wrong & Right & Accuracy (\%) & Class & Wrong & Right & Accuracy (\%) \\
\hline
\CC  & ~1  & 19 & ~95     & \OO  & 13 & ~7  & ~35 \\
\CR & ~8  & 12 & ~60     &    \UU  & ~0  & 20 & 100 \\
\LL  & ~0  & 20 & 100    &   \UR & ~4  & 16 & ~80 \\
\LR & ~0  & 20 & 100     &    \YY  & ~0  & 20 & 100 \\
\hline
&&&& \bf Overall & 26 & \!\!134 & \!\!83.75 
\end{tabular}
\end{table}

Table \ref{tab:backbone_baselines} summarizes the performance of the evaluated architectures. We additionally adapt the NetLay multi-label encoding~\cite{gogawale2024netlay} 
to our 8-class task (Table~\ref{tab:backbone_baselines}, right); the best result reaches only 52\% accuracy and 0.49 macro-$F_1$ without layout-preserving augmentations, confirming that the separator-aware 
training strategy is the decisive contribution. ConvNeXt-Tiny performs substantially better than the others, achieving an accuracy of 0.90 and a macro-$F_1$ of 0.88, substantially outperforming EfficientNet-B0 and VGG16 (acc.\@ 0.45, macro-$F_1$ 0.41), as well as ResNet-18 and ViT-B/16 (macro-$F_1$ $\le$ 0.17). On a per-class basis, ConvNeXt achieves perfect or near-perfect performance on most layouts (e.g.\@ \CR/\LR/\UU), with remaining errors concentrated in the rarest and most ambiguous classes (\YY and \UR), suggesting confusion among separator patterns when examples are extremely limited.
In low-resource regimes, ViTs often underperform unless heavily regularized or trained on large datasets.
ResNet-18 suffers a similar fate, likely due to its smaller receptive field and aggressive spatial downsampling, which can weaken the thin, delicate separator lines crucial for identifying medieval layout classes. The strong performance of ConvNeXt-Tiny supports our data-scarcity hypothesis: with limited samples, the model cannot rely on memorizing textures. Instead, CNN inductive biases encourage learning contiguous edges, while the ViT-inspired design helps integrate these cues into coherent global layout structures.
The only notable struggle for ConvNeXt-Tiny is the \YY class (recall of 0.33, $F_1= 0.50$), which represents the most intricate layout with central text flanked by multiple distinct elements. 
This slight degradation highlights that, while ConvNeXt effectively models global structure from minimal data, highly fragmented layouts with inner and outer separators remain challenging  for zero-to-few-shot historical document analysis.

\paragraph{Evaluation on a  blind corpus.}
To examine how the classifier behaves outside our CLC dataset, we conducted a blind-corpus sanity check on a large corpus drawn from the book 
%\todo{can't mention Midrash until deanonymized} 
collections of the National Library of Israel (NLI). These  were not used for training, validation, or hyperparameter tuning.
We emphasize that this is not a fully labeled out-of-domain benchmark; rather, it is a precision-oriented stress test on high-confidence predictions in a large unlabeled archive.
Inference was performed over a directory containing 36,215 book folders. For each folder, up to 10 page images were randomly sampled (or fewer when unavailable), yielding a maximum of 362,150 candidate pages.
The trained classifier model (Full configuration: Binary+Flips+Masking) was applied without any fine-tuning. To focus on highly confident predictions, we retained only outputs whose maximum softmax probability exceeded $0.99$.
This procedure identified:
 3,932 book folders containing at least one page classified as complex;
and 9,339 page images satisfying the confidence threshold.

Because the corpus lacks ground truth layout annotations, we conducted a manual visual evaluation on a stratified sample of the high-confidence predictions. For each of the eight  classes, 20 images were randomly selected.
%, resulting in 160 images in total.
%
Each image was visually inspected and assigned its true layout class, which was then compared to the model prediction. See Table~\ref{tab:external_blind}. Out of 160 inspected samples, 134 were correct, giving an overall accuracy of 83.75\% on this verified subset. Performance was lower for \CR and especially for \OO layouts.
%(Fig.~\ref{fig:blind_false_pos_grid}).

\begin{ignore}
    
\begin{figure*}[t]
\centering
%\setlength{\tabcolsep}{2pt}
%\renewcommand{\arraystretch}{0}

% top-aligned fixed-width container
\newcommand{\imgtop}[1]{%
\vtop{%
  \hbox{%
    \parbox[t]{0.11\textwidth}{%
      \centering
      \includegraphics[width=\linewidth,keepaspectratio]{#1}%
    }%
  }%
}%
}

\begin{tabular}{cccccccc}
\textbf{C} & \textbf{C2} & \textbf{L} & \textbf{L2} & \textbf{O} & \textbf{U} & \textbf{U2} & \textbf{Y} \\[2pt]

% Row 1
\imgtop{figures/blind_test/true_pos/c_2.jpg} &
\imgtop{figures/blind_test/true_pos/c2_2.jpg} &
\imgtop{figures/blind_test/true_pos/l_2.jpg} &
\imgtop{figures/blind_test/true_pos/l2_1.jpg} &
\imgtop{figures/blind_test/true_pos/o_8.jpg} &
\imgtop{figures/blind_test/true_pos/u_3.jpg} &
\imgtop{figures/blind_test/true_pos/u2_4.jpg} &
\imgtop{figures/blind_test/true_pos/y_4.jpg} \\[2pt]

% Row 2
\imgtop{figures/blind_test/true_pos/c_9.jpg} &
\imgtop{figures/blind_test/true_pos/c2_7.jpg} &
\imgtop{figures/blind_test/true_pos/l_11.jpg} &
\imgtop{figures/blind_test/true_pos/l2_2.jpg} &
\imgtop{figures/blind_test/true_pos/o_9.jpg} &
\imgtop{figures/blind_test/true_pos/u_7.jpg} &
\imgtop{figures/blind_test/true_pos/u2_5.jpg} &
\imgtop{figures/blind_test/true_pos/y_10.jpg} \\[2pt]

% Row 3
\imgtop{figures/blind_test/true_pos/c_11.jpg} &
\imgtop{figures/blind_test/true_pos/c2_10.jpg} &
\imgtop{figures/blind_test/true_pos/l_13.jpg} &
\imgtop{figures/blind_test/true_pos/l2_3.jpg} &
\imgtop{figures/blind_test/true_pos/o_14.jpg} &
\imgtop{figures/blind_test/true_pos/u_15.jpg} &
\imgtop{figures/blind_test/true_pos/u2_10.jpg} &
\imgtop{figures/blind_test/true_pos/y_13.jpg} \\[2pt]

% Row 4
\imgtop{figures/blind_test/true_pos/c_17.jpg} &
\imgtop{figures/blind_test/true_pos/c2_18.jpg} &
\imgtop{figures/blind_test/true_pos/l_18.jpg} &
\imgtop{figures/blind_test/true_pos/l2_4.jpg} &
\imgtop{figures/blind_test/true_pos/o_20.jpg} &
\imgtop{figures/blind_test/true_pos/u_20.jpg} &
\imgtop{figures/blind_test/true_pos/u2_11.jpg} &
\imgtop{figures/blind_test/true_pos/y_17.jpg} \\

\end{tabular}
\caption{
Correctly classified examples from the 
%external Midrash 
test corpus.}
\label{fig:blind_true_pos_grid}
\end{figure*}
\end{ignore}

\newcolumntype{C}{>{\centering\arraybackslash}m{0.11\textwidth}}
\begin{ignore}\begin{figure}[t]
\centering

\newcommand{\imgtop}[1]{%
\vtop{%
  \hbox{%
    \parbox[t]{0.14\textwidth}{%
      \centering
      \includegraphics[height=20mm]{#1}%
    }%
  }%
}%
}
\resizebox{!}{20mm}{
\begin{tabular}{CCCCCC}
\imgtop{figures/blind_test/false_pos/c2_3.jpg} &
\imgtop{figures/blind_test/false_pos/c2_4.jpg} &
\imgtop{figures/blind_test/false_pos/c2_8.jpg} &
\imgtop{figures/blind_test/false_pos/c2_9.jpg} &
\imgtop{figures/blind_test/false_pos/o_2.jpg} &
\imgtop{figures/blind_test/false_pos/o_4.jpg} \\
\CR &\CR &\CR &\CR & \OO &\OO\\
\imgtop{figures/blind_test/false_pos/o_7.jpg} &
\imgtop{figures/blind_test/false_pos/o_7.jpg} & 
\imgtop{figures/blind_test/false_pos/u2_1.jpg} &
\imgtop{figures/blind_test/false_pos/u2_6.jpg} &
\imgtop{figures/blind_test/false_pos/u2_8.jpg} &
\imgtop{figures/blind_test/false_pos/u2_9.jpg} \\
\OO & \OO & \UR &\UR &\UR &\UR 
\end{tabular}
}
\caption{
misclassified examples from the %external Midrash 
test corpus.}
\label{fig:blind_false_pos_grid}
\end{figure}
\end{ignore}

\paragraph{Ablation.}
To systematically quantify the contributions of our proposed layout-preserving augmentations, we conducted an extensive ablation study using the best-performing ConvNeXt-Tiny architecture. All experiments are evaluated on the same 8-way test set (31 pages). 

\paragraph{Ablation Settings.}
We consider (i) \textbf{Binary only}, which uses only binarized inputs without geometric augmentations; (ii) \textbf{Binary+Flips}, which additionally applies horizontal/vertical reflections with label-consistent remapping for asymmetric layouts; (iii) \textbf{Binary+Masking}, which uses only the proposed narrow anisotropic Gaussian masking to suppress incidental text while preserving separator structure; (iv) \textbf{Full}, which combines Binary+Flips+Masking; and (v) \textbf{Mask strength} sweeps, where we vary a masking hyperparameter $\eta \in \{1,3,5\}$ controlling the strength of  anisotropic suppression.

\paragraph{Quantitative results.}
Table~\ref{tab:ablations_main_eta} shows that Binary preprocessing alone performs poorly (acc.\@ 0.32, macro-$F_1$ 0.29), indicating that simple foreground/background normalization is insufficient to learn separator-defined global geometry from tiny supervision. 
Notably, classes with nested semantic regions (\LR and \YY) completely fail to converge ($F_1$: 0.00) in this low-resource regime.
Reflection-based augmentation in the \textbf{Binary+Flips} strategy, together with a fixed label-transformation mapping, improves  accuracy to 0.45.  While this successfully expands the dataset and enforces orientation invariance (boosting \CC's $F_1$ from 0.44 to 0.62), the model remains heavily distracted by local textual textures, resulting in marginal gains for complex asymmetric layouts. We also evaluate a \textbf{Binary+Standard Augmentations}  baseline using common transforms (random rotation, affine, perspective,  and Gaussian blur), which achieves 0.68 accuracy and 0.63 macro-$F_1$. While this exceeds Binary+Flips, the full pipeline with separator-preserving 
masking and label-consistent reflections yields substantially stronger 
performance, confirming that task-aware augmentation design matters 
beyond generic regularization.

\begin{table}[t]
\caption{Ablation results on CLC using ConvNeXt-Tiny (31 pages), showing \textbf{Acc}uracy and macro-averaged \textbf{P}recision, \textbf{R}ecall, and \textbf{$F_1$}. 
%Higher is better. 
Best in \textbf{bold}.}
\label{tab:ablations_main_eta}\label{tab:ablations_components}\label{tab:ablations_eta}
\qquad
\begin{minipage}{0.55\linewidth}
%\centering
%\small
%\captionof{table}{Component ablations.}
%\label{tab:ablations_components}
\begin{tabular}{lcccc}
\toprule
\textbf{Config} & \textbf{Acc.} & \textbf{P} & \textbf{R} & \textbf{$F_1$} \\
\hline
Binary only                     & 0.32 & 0.34 & 0.31 & 0.29 \\
Binary+Flips                  & 0.45 & 0.44 & 0.47 & 0.42 \\
Binary+Masking                & 0.68 & 0.69 & 0.61 & 0.59 \\
Binary+Std.\ Aug. & 0.68 & 0.76 & 0.62 & 0.63 \\
\textbf{Full} (Masking+Flips) & \textbf{0.90} & \textbf{0.93} & \textbf{0.88} & \textbf{0.88} 
\end{tabular}
\end{minipage}
\quad
\begin{minipage}{0.4\linewidth}
%\centering
%\small
%\captionof{table}{Mask strength sweep ($\eta$).}
%\label{tab:ablations_eta}
\vspace*{-9pt}
\begin{tabular}{lcccc}
\toprule
\textbf{$\eta$} & \textbf{Acc.} & \textbf{P} & \textbf{R} & \textbf{$F_1$} \\
\hline
$\eta{=}1$ & 0.55 & 0.56 & 0.53 & 0.52 \\
$\eta{=}3$ & 0.71 & 0.82 & 0.70 & 0.69 \\
$\eta{=}5$ & 0.81 & 0.85 & 0.79 & 0.79 
\end{tabular}
\end{minipage}
\end{table}

In contrast, \textbf{Binary+Masking} produces a substantial jump (acc.\@ 0.68, macro-$F_1$ 0.59), 
supporting the hypothesis that the task is fundamentally \emph{separator-driven}.
By stochastically suppressing high-frequency textual components while preserving the contiguous geometric paths of separators, the masking forces the CNN’s inductive bias to focus strictly on layout-defining topological boundaries. 
Hence, the model is compelled to base its predictions on global geometric structures, such as long separator strokes, block boundaries, and their spatial arrangement, rather than local content- or style-specific patterns.

Finally, combining all components (\textbf{Full}) achieves the best overall performance (acc.\@ 0.90, macro-$F_1$ 0.88), indicating that flips and masking are complementary: flips increase effective sample diversity and promote invariance, while masking acts as a structure-preserving regularizer that improves generalization.

\paragraph{Effect of masking strength.}
Varying the masking hyperparameter highlights a clear sensitivity: weaker masking ($\eta{=}1$) underperforms (acc.\@ 0.55, macro-$F_1$ 0.52), while moderate-to-strong masking improves robustness ($\eta{=}3$: acc.\@ 0.71, macro-$F_1$ 0.69; $\eta{=}5$: acc.\@ 0.81, macro-$F_1$ 0.79). This trend suggests that sufficiently strong suppression is required to consistently remove text cues; otherwise, the model partially relies on residual texture signals that do not transfer across pages. This directional filtering acts as a line-tracking mechanism: it aggressively blurs orthogonal high-frequency noise (text) while constructively reinforcing continuous low-frequency paths (separators).
Nevertheless, the best performance is obtained by the \textbf{Full} configuration, implying that masking alone is not a substitute for geometric augmentation and label-consistent symmetry handling.

% \subsection{Minimizing Annotation Cost via Active Learning}
% \label{subsec:results_al}

% While our full augmentation pipeline establishes strong generalization under data scarcity, annotation in historical  collections remains expensive and slow. 
% We therefore evaluate whether an active learning workflow can  reduce the number of labeled pages required to reach competitive performance. 

% As shown in Fig.~\ref{fig:both}(left), separator-preserving augmentation substantially improves performance across labeling budgets. Fig.~\ref{fig:both}(right) further illustrates that entropy-based acquisition improves label-efficiency in the mid-budget regime.

\paragraph{Minimizing cost via active learning.}
\label{subsec:results_al}
Annotation remains a bottleneck in historical document analysis, since complex
layouts often require domain-aware inspection. We therefore evaluate whether
active learning can improve label efficiency when expanding CLC.

\begin{figure}[t]
\centering
\includegraphics[
    width=0.5\linewidth,
    trim=0 0 0 0.9cm,
    clip
]{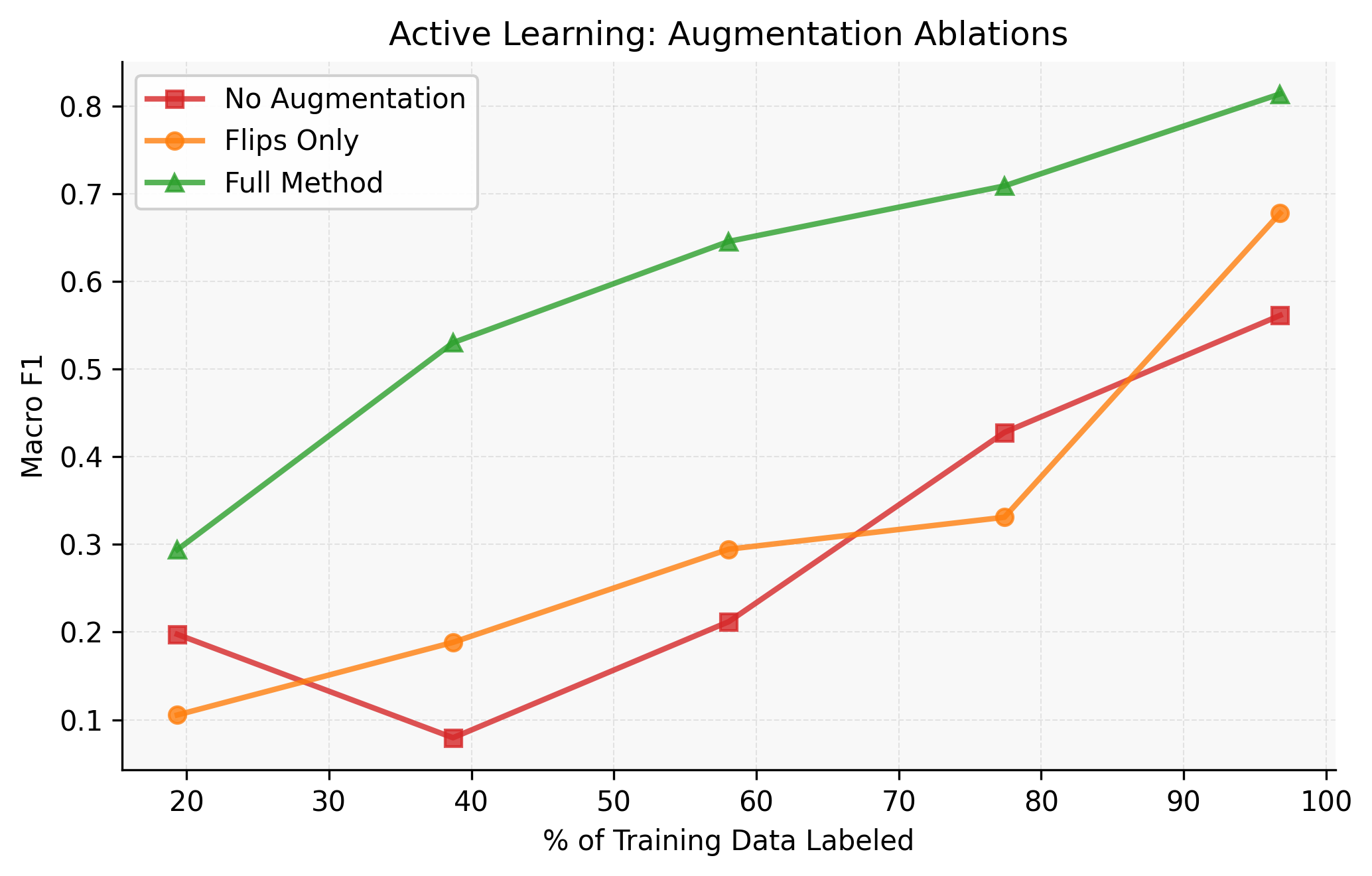}
\caption{
Active learning under label scarcity. Macro-$F_1$ learning curves are shown
for three augmentation settings, demonstrating that the full
layout-preserving pipeline improves performance throughout the low-label
regime.
}
\label{fig}
\end{figure}

Figure~\ref{fig} shows that the dominant factor in label-efficient learning is the augmentation strategy. Under entropy acquisition, the full layout-preserving pipeline achieves a macro-$F_1$ AULC of 0.61, compared with 0.30 for Binary+Flips and 0.28 for Binary only. This large shift indicates that separator-preserving masking and label-consistent reflections improve learning consistently across all labeling budgets, rather than only at the final training phase.

\rev{Evaluating the acquisition rules reveals a secondary effect relative to the augmentation strategy.} On accuracy, the AULC is identical for Random, Entropy, and Margin acquisition (0.65), with all three converging to the same final accuracy of 0.87. However, macro-$F_1$ is more sensitive to the acquisition choice: Margin yields the highest AULC (0.62), compared with 0.61 for both Entropy and Random. \rev{Its advantage appears mainly in the mid-budget regime: at 60\% labeled data, Margin reaches a macro-$F_1$ of 0.69 (compared to 0.59 for Random), and at 80\% it reaches 0.76 (compared to 0.73 for Random). At the earliest post-initial budget, however, Margin slightly trails Random, suggesting that acquisition-rule differences are less reliable when the labeled set is extremely small. Overall, these results suggest that the primary label-efficiency gain comes from the separator-aware augmentation pipeline, while margin-based acquisition provides a lightweight mechanism for prioritizing structurally informative pages during dataset expansion.}

\section{Conclusion}

We addressed the challenging task of complex-layout classification in low-resource historical documents, motivated by the practical need to route heterogeneous documents to layout-aware transcription and segmentation pipelines. \rev{To support controlled evaluation under annotation scarcity, we curated the CLC dataset, consisting of 155 high-resolution pages spanning eight separator-defined layout categories.}
We introduced layout-preserving augmentations that suppress incidental textual detail while retaining separator structure, and combined them with label-consistent reflection transformations to expand the effective training distribution without introducing label noise. Masking, combined with label-consistent spatial reflections, achieved 90\% accuracy on our dataset. Furthermore, an external blind evaluation on the massive NLI corpus demonstrated strong generalization behavior yielding an 83.75\% accuracy \rev{on a manually verified high-confidence subset, suggesting that the classifier can generalize beyond the curated dataset. Finally, our preliminary active-learning study shows that the augmentation pipeline is the primary driver of label efficiency, while margin-based acquisition provides a lightweight strategy for prioritizing structurally informative pages during dataset expansion. Future work will extend the approach to handwritten manuscript pages and integrate script-type recognition for layout-aware OCR routing.}

%Future work includes learning to classify complex layouts of handwritten manuscript pages, combined with automatic recognition of script types and modes (there are fifteen for medieval manuscripts in general). This will lead to much better OCR for manuscripts and printed books, and will allow us to approach historical documents with more confidence and accuracy.  

\subsubsection*{Acknowledgments.}
This research was funded in part by the European Union as part of  "MiDRASH" [\url{https://www.midrash.eu}] (ERC project no.\@ 101071829, with 
principal investigators: Nachum Dershowitz, Tel Aviv University; Judith Olszowy-Schlanger, EPHE-PSL; Avi Shmidman, Bar-Ilan University; and Daniel Stökl Ben Ezra, EPHE-PSL).
Views and opinions expressed are, however, those of the authors only and do not necessarily reflect those of the European Union or the European Research Council Executive Agency. Neither the European Union nor the granting authority can be held responsible for them.

%\todo{open up abbreviations like CVPR}
\bibliographystyle{splncs04}
\bibliography{zbib}

@INPROCEEDINGS{dla,
  author={Zhang, Chunhu and Ibrayim, Mayire and Hamdulla, Askar},
  OPTbooktitle={Intl. Conf. on Virtual Reality, Human-Computer Interaction and Artificial Intelligence (VRHCIAI)}, 
    booktitle={VRHCIAI}, 
  title={A Methodological Study of Document Layout Analysis}, 
  year={2022},
  volume={},
  number={},
  pages={12--17},
  OPTdoi={10.1109/VRHCIAI57205.2022.00009}}

@INPROCEEDINGS{dla2,
  author={Lee, Joonho and Hayashi, Hideaki and Ohyama, Wataru and Uchida, Seiichi},
  OPTbooktitle={Intl. Conf. on Document Analysis and Recognition (ICDAR)},   booktitle={ICDAR}, 
  title={Page Segmentation using a Convolutional Neural Network with Trainable Co-occurrence Features}, 
  year={2019},
  volume={},
  number={},
  pages={1023--1028},
  OPTdoi={10.1109/ICDAR.2019.00167}}

@inproceedings{dla3,
  title={Learning to Extract Semantic Structure from Documents Using Multimodal Fully Convolutional Neural Networks},
  author={Xiao Yang and Ersin Yumer and Paul Asente and Mike Kraley and Daniel Kifer and C. Lee Giles},
  OPTbooktitle={IEEE Conf. on Computer Vision and Pattern Recognition (CVPR)},
  booktitle={CVPR},
 year={2017},
  pages={4342--4351},
  OPTurl={https://api.semanticscholar.org/CorpusID:2272015}
}

@inproceedings{publaynet,
  title={{PubLayNet}: largest dataset ever for document layout analysis},
  author={Zhong, Xu and Tang, Jianbin and Yepes, Antonio Jimeno},
  OPTbooktitle={Intl. Conf. Document Analysis and Recognition},
  booktitle="ICDAR",
  year={2019},
  volume={},
  number={},
  pages={1015--1022},
  OPTdoi={10.1109/ICDAR.2019.00166},
  ISSN={1520-5363},
  month=sep,
  OPTorganization={IEEE}
}

@inproceedings{doclaynet2022,
  title = {{DocLayNet}: A Large Human-Annotated Dataset for Document-Layout Segmentation},
  OPTdoi = {10.1145/3534678.353904},
  OPTurl = {https://doi.org/10.1145/3534678.3539043},
  author = {Pfitzmann, Birgit and Auer, Christoph and Dolfi, Michele and Nassar, Ahmed S and Staar, Peter W J},
  year = {2022},
  isbn = {9781450393850},
  OPTpublisher = {Association for Computing Machinery},
  OPTaddress = {New York},
  OPTbooktitle = {Proc. 28th ACM SIGKDD Conf. on Knowledge Discovery and Data Mining},
  pages = {3743--3751},
  numpages = {9},
  location = {Washington DC},
  booktitle = {KDD {'}22}
}

@article{udiads,
author = {Zottin, Silvia and De Nardin, Axel and Colombi, Emanuela and Piciarelli, Claudio and Pavan, Filippo and Foresti, Gian Luca},
title = {{U-DIADS-Bib}: A full and few-shot pixel-precise dataset for document layout analysis of ancient manuscripts},
year = {2024},
OPTissue_date = {Jul 2024},
publisher = {Springer},
address = {Berlin},
volume = {36},
number = {20},
issn = {0941-0643},
OPTurl = {https://doi.org/10.1007/s00521-023-09356-5},
doi = {10.1007/s00521-023-09356-5},
journal = {Neural Comput. Appl.},
month = jan,
pages = {11777--11789},
numpages = {13}
}

@InProceedings{fest,
author="Zottin, Silvia
and De Nardin, Axel
and Branca, Giuseppe
and Piciarelli, Claudio
and Foresti, Gian Luca",
OPTeditor="Yin, Xu-Cheng
and Karatzas, Dimosthenis
and Lopresti, Daniel",
title="{ICDAR} 2025 Competition on FEw-Shot Text Line Segmentation of Ancient Handwritten Documents {(FEST)}",
OPTbooktitle="Document Analysis and Recognition --  ICDAR 2025",
year="2026",
booktitle="ICDAR",
OPTpublisher="Springer Nature Switzerland",
address="Cham",
pages="586--602",
isbn="978-3-032-04630-7"
}

@Article{brazil,
AUTHOR = {Santos Júnior, Eder Silva dos and Paixão, Thuanne and Alvarez, Ana Beatriz},
TITLE = {Comparative Performance of {YOLOv8}, {YOLOv9}, {YOLOv10}, and {YOLOv11} for Layout Analysis of Historical Documents Images},
JOURNAL = {Applied Sciences},
VOLUME = {15},
YEAR = {2025},
NUMBER = {6},
ARTICLE-NUMBER = {3164},
OPTurl = {https://www.mdpi.com/2076-3417/15/6/3164},
ISSN = {2076-3417},
OPTdoi = {10.3390/app15063164}
}

@InProceedings{m6doc,
    author    = {Cheng, Hiuyi and Zhang, Peirong and Wu, Sihang and Zhang, Jiaxin and Zhu, Qiyuan and Xie, Zecheng and Li, Jing and Ding, Kai and Jin, Lianwen},
    title     = {{M$^6$Doc}: A Large-Scale Multi-Format, Multi-Type, Multi-Layout, Multi-Language, Multi-Annotation Category Dataset for Modern Document Layout Analysis},
    OPTbooktitle = {Proc. IEEE/CVF Conf. on Computer Vision and Pattern Recognition (CVPR)},
    booktitle={CVPR},
    month     = jun,
    year      = {2023},
    pages     = {15138--15147}
}

@INPROCEEDINGS{prima,
  author={Antonacopoulos, Apostolos and Bridson, David and Papadopoulos, Christos and Pletschacher, Stefan},
  OPTbooktitle={10th Intl. Conf. on Document Analysis and Recognition}, 
  booktitle="ICDAR",
  title={A Realistic Dataset for Performance Evaluation of Document Layout Analysis}, 
  year={2009},
  volume={},
  number={},
  pages={296--300},
  OPTdoi={10.1109/ICDAR.2009.271}}

@INPROCEEDINGS{diva,
  author={Simistira, Foteini and Seuret, Mathias and Eichenberger, Nicole and Garz, Angelika and Liwicki, Marcus and Ingold, Rolf},
  OPTbooktitle={15th Intl. Conf. on Frontiers in Handwriting Recognition}, 
 booktitle={ICFHR}, 
  title={{DIVA-HisDB}: A Precisely Annotated Large Dataset of Challenging Medieval Manuscripts}, 
  year={2016},
  volume={},
  number={},
  pages={471--476},
  OPTdoi={10.1109/ICFHR.2016.0093}}

@inproceedings{scut,
  title={{SCUT-CAB}: A New Benchmark Dataset of Ancient {Chinese} Books with Complex Layouts for Document Layout Analysis},
  author={Hiuyi Cheng and Cheng Jian and Sihang Wu and Lianwen Jin},
  OPTbooktitle={Intl. Conf. on Frontiers of Handwriting Recognition (ICFHR)},
  booktitle={ICFHR},
  year={2022}
}

@inproceedings{hjd,
  title={A Large Dataset of Historical {Japanese} Documents with Complex Layouts},
  author={Shen, Zejiang and Zhang, Kaixuan and Dell, Melissa},
year = {2020},
month = jun,
booktitle="CVPR Workshops",
pages = {2336--2343},
OPTdoi = {10.1109/CVPRW50498.2020.00282}
}

@inproceedings{feature,
author = {Josi, Frieda and Wartena, Christian and Heid, Ulrich},
year = {2022},
month = jan,
volume = 12,
pages = {},
title = {Preparing Legal Documents for {NLP} Analysis: Improving the Classification of Text Elements by Using Page Features},
booktitle="8th Intl. Conf. on Natural Language Processing (NATP)",
OPTdoi = {10.5121/csit.2022.120102}
}

@inproceedings{news,
author = {Hamdi, Ahmed and Linhares Pontes, Elvys and Boros, Emanuela and Nguyen, Thi Tuyet Hai and Hackl, G\"{u}nter and Moreno, Jose G. and Doucet, Antoine},
title = {A Multilingual Dataset for Named Entity Recognition, Entity Linking and Stance Detection in Historical Newspapers},
year = {2021},
isbn = {9781450380379},
OPTpublisher = {Association for Computing Machinery},
OPTaddress = {New York},
OPTurl = {https://doi.org/10.1145/3404835.3463255},
OPTdoi = {10.1145/3404835.3463255},
OPTbooktitle = {Proc. 44th Intl. ACM SIGIR Conf. on Research and Development in Information Retrieval},
pages = {2328--2334},
numpages = {7},
location = {Virtual Event, Canada},
booktitle = {SIGIR {'}21}
}

@inproceedings{graphical,
  title={Graphical Object Detection in Document Images},
  author={Saha, Ranajit and Mondal, Ajoy and Jawahar, C. V.},
  OPTbooktitle={Intl. Conf. on Document Analysis and Recognition (ICDAR)},
  booktitle ="ICDAR",
  pages={51--58},
  year={2019},
  OPTorganization={IEEE}
}

@article{gogawale2024netlay,
  title={{NetLay}: Layout Classification Dataset for Enhancing Layout Analysis},
  author={Gogawale, Sharva and Bambaci, Luigi and Kurar-Barakat, Berat and Vasyutinsky Shapira, Daria  and St{\"o}kl Ben Ezra, Daniel  and Dershowitz, Nachum},
  journal={Magaz{\'e}n: Intl. J. for Digital and Public Humanities},
  year={2024},
   volume= 5, OPTnumber= 2, pages="1--14"
}

@inproceedings{convnext,
  author  = {Zhuang Liu and Hanzi Mao and Chao-Yuan Wu and Christoph Feichtenhofer and Trevor Darrell and Saining Xie},
  title   = {A {ConvNet} for the 2020s},
  OPTbooktitle = {Proc. IEEE/CVF Conf. on Computer Vision and Pattern Recognition (CVPR)},
    booktitle = {CVPR},
  year    = {2022},
}

@article{kraken,
author="Stokes, P. and B. Kiessling and D. {Stökl Ben Ezra} and R. Tissot and E. H. Gargem",
title="The {eScriptorium} {VRE} for Manuscript Cultures",
journal="Classics@ Journal",
OPTjournal="Classics@ Journal: Ancient Manuscripts and Virtual Research Environments", 
volume=18, year=2021,
OPTurl="https://classics-at.chs.harvard.edu/classics18-stokes-kiessling-stokl-ben-ezra-tissot-gargem"}

\end{document}